\documentclass{article}

\PassOptionsToPackage{numbers}{natbib}
\usepackage{natbib}
\usepackage[nonatbib, preprint]{neurips_2022}
\usepackage[utf8]{inputenc} % allow utf-8a input
\usepackage[T1]{fontenc}    % use 8-bit T1 fonts
\usepackage{hyperref}       % hyperlinks

\usepackage{url}            % simple URL typesetting
\usepackage{booktabs}       % professional-quality tables
\usepackage{amsfonts}       % blackboard math symbols
\usepackage{nicefrac}       % compact symbols for 1/2, etc.
\usepackage{microtype}      % microtypography
\usepackage{xcolor}         % colors
\usepackage{graphicx}
\usepackage{amsmath}
\usepackage{amssymb}        % complement notation
\usepackage{algorithm}
\usepackage[noend]{algpseudocode}
\usepackage{multirow}
\usepackage{commath}
\usepackage{subcaption}
\usepackage{tabularx}
\usepackage{makecell}
\usepackage{etoc}
\usepackage{comment}
\usepackage{cuted}
\usepackage{caption}

\newcommand{\E}{\textbf{E}}
\newcommand{\bb}[1]{\mathbb{#1}}
\newcommand{\obar}[1]{\mkern 1.5mu\overline{\mkern-1.5mu#1\mkern-1.5mu}\mkern 1.5mu}

\newtheorem{theorem}{Theorem}
\newtheorem{lemma}{Lemma}

\newtheorem{assumption}{Assumption}
\newtheorem{definition}{Definition}

\DeclareMathOperator*{\argmin}{arg\,min}

\title{Testing for context-dependent changes in neural encoding in naturalistic experiments}

\author{%
  Yenho Chen\\
  Machine Learning Center\\
  Georgia Tech University\\
  Atlanta, GA 30308  \\
  \texttt{yenho@gatech.edu} \\
  \And
  Carl W. ~Harris \\
  Machine Learning Team\\
  National Institute of Mental Health \\
  Bethesda, MD 20892 \\
  \texttt{carl.harris@nih.gov} \\
  \And
  Xiaoyu Ma \\
  Section on Synapse Development Plasticity\\
  National Institute of Mental Health \\
  Bethesda, MD 20892 \\
  \texttt{xiaoyu.ma@nih.gov} \\
  \And
  Zheng Li \\
  Section on Synapse Development Plasticity\\
  National Institute of Mental Health\\
  Bethesda, Maryland 20892 \\
  \texttt{zheng.li@nih.gov} \\
  \And
  Francisco ~Pereira\\
  Machine Learning Team\\
  National Institute of Mental Health\\
  Bethesda, Maryland 20892 \\
  \texttt{francisco.pereira@nih.gov} \\
  \And
  Charles Y.~Zheng\\
  Machine Learning Team\\
  National Institute of Mental Health\\
  Bethesda, Maryland 20892 \\
  \texttt{charles.zheng@nih.gov} \\
}

\begin{document}

\maketitle

\begin{abstract}
We propose a decoding-based approach to detect context effects on neural codes in longitudinal neural recording data. The approach is agnostic to how information is encoded in neural activity, and can control for a variety of possible confounding factors present in the data. We demonstrate our approach by determining whether it is possible to decode location encoding from prefrontal cortex in the mouse and, further, testing whether the encoding changes due to task engagement.
\end{abstract}

\section{Introduction and Related Work}

If we accept the premise of Douglas Adam's \emph{Hitchhiker's Guide to the Galaxy}, then mice are actually the most intelligent species on the planet and delight in manipulating 
%the human
researchers studying them. One aspect of their success may be the increasing popularity of naturalistic experiments. These %experiments
have mice engaging in goal-directed behavior, and determining 
when and how to act in face of their environment.
%the timing and selection of their actions. 
Despite the importance of naturalistic tasks  for establishing the ecological validity of neuroscientific findings, they break many assumptions required of classical statistical approaches, such as balanced sampling, and thus require new methodology capable of controlling multiple layers of confounding factors. 
The motivation for the work described in this article is to analyze an experiment where mice are required to maintain spatial location information during the performance of a task. The experimental question is whether that information is encoded in mouse prefrontal cortex and, if so, whether this encoding is modulated by goal-oriented behavior. This is a pertinent question
%, as many recent works have investigated and shown
given the known effects of various factors on encoding, e.g. goals modulate auditory encodings in A1 \cite{fritz_rapid_2003}, and stimulus uncertainty affects encoding in the premotor dorsal cortex \cite{glaser_population_2018}. The classical approach to detecting such changes in neural code is to test individual neurons for changes in their tuning curves between contexts, via two-sample univariate distribution tests or general linear mixed effect models \cite{reed_statistical_2010}. However, improved power can be obtained by testing for multivariate effects between groups of neurons, using Hotelling's $T^2$ test or other two-sample tests \cite{panda2019nonparametric}.  Yet another category of approaches can be found in the distribution shift literature \cite{moreno-torres_unifying_2012, rabanser2019failing}, which is concerned with testing for differences in the distribution of covariates between samples, under the assumption that the conditional probability of the label given the data, $p(y|x)$, is the same in both samples. In this formulation, testing for changes in encoding between goal and non-goal oriented tasks is an equivalent problem.  %It is important, therefore, not to apply methods that are designed for testing changes in the distribution $p(y|x)$ (e.g. \cite{lipton2018detecting}), as these test an effect in the wrong causal direction.

In this paper, we introduce a novel decoding-based methodology, using cross-context decoding accuracies \cite{kaplan_multivariate_2015}
to test for context-dependent changes in neural code under very general data sampling assumptions. In this paradigm, a subset of the data from one context is used to train a classifier. The classifier is then evaluated on a second, held-out set from the same context. If the classifier's performance on held-out data from the same context as the training data is significantly better than its performance on data from a different context, there is evidence that that that the information contained in each context differs. Decoder accuracies are commonly used in the neuroscience literature, particularly in neuroimaging, as reviewed by \cite{kaplan_multivariate_2015}, both because of improved capacity to incorporate multivariate and nonlinear encodings, but also because decoding accuracies are highly interpretable (in contrast to information theoretic quantities or statistical quantities such as effect size). However, existing studies after fail to account for the myriad potential confounds endemic to naturalistic experiments, motivating the current study. 

Here, we provide a general tool for isolating environmental and cognitive factors that affect how neurons encode information, based on a bidirectional measure of cross-context accuracy (i.e., one that incorporates cross-classification accuracies from context A $\rightarrow$ context B \textit{and} context B $\rightarrow$ context A). We demonstrate the utility of our model using a specific naturalistic task dataset and linear naïve Bayes classifier, and note its applications to a wide variety of neuroscientific applications, across different animals, types of stimuli (e.g., auditory, olfactory, social, etc.), context effects (e.g., learning, social effects, physiological condition), and with different types of classifiers (e.g., support vector machines, neural networks, etc.). We provide a statistically principled control against false positives, while controlling for a variety of confounding factors:  variation in the amount of data, variation in the target distribution, short-range temporal dependence in the neural signal, and additional measured confounders.  Thus, our approach is more comprehensive than the synthetic null data generation methods introduced by \cite{elsayed_structure_2017}, which account for temporal correlations but not for label imbalance.  We use a combination of existing and novel methodologies, integrating all of them within a pipeline where specific subroutines can be included as needed.  
Using our approach, we infer that location encoding differs between sessions in which a mouse has been trained to seek a reward and those in which it moves freely in a maze.  Furthermore, in simulations, we show that our approach is more conservative in a wider range of experimental settings than existing univariate and multivariate tests.
% We note also that our method for inferring classifier generalization accuracy from performance on correlated time series data may be interesting to the wider machine learning literature. 
%We show that our approach is feasible on our mouse data. Specifically, we demonstrate that location encoding differs between a task context for which the mouse has been trained to seek a reward and a free-running context where the mouse moves freely in the maze.

\section{Data}

Neural time series were recorded at a 30k Hz sampling rate with a 64-channel microdrive tetrode array implanted in the prefrontal cortex region of the mouse brain (AC 1.8mm, ML 0.4mm, DV 1.8mm). Covariates such as head orientation and mouse location were extracted from video footage captured at 25fps from an overhead perspective using Topscan Suite 3.0 (Clever Sys) software. Data were continuously recorded as a single time series throughout the experiment. Spike waveforms were detected from 250 Hz high-pass filtered wideband signals by a threshold at $-3.5 \sigma$, and then checked with PCA, in Offline Sorter (Plexon). Neural firings were converted into spike counts during 40-ms intervals, to match the sampling rate of the video camera.

% \begin{strip}
% \centering
%   \includegraphics[width=0.8\textwidth]{decodingMaze-cropped.png}

%   \captionof{figure}{\textbf{A)} T-maze layout and discretization scheme. The holding zone is located at the base (2) while water spouts are placed in the arms (0). Dashed lines represent doors that can be used to constrain mouse movement. Location labels are symmetric between left and right arms.}
%   \label{fig:t-maze}
% \end{strip}

\begin{figure*}[ht]
\includegraphics[width=0.90\textwidth]{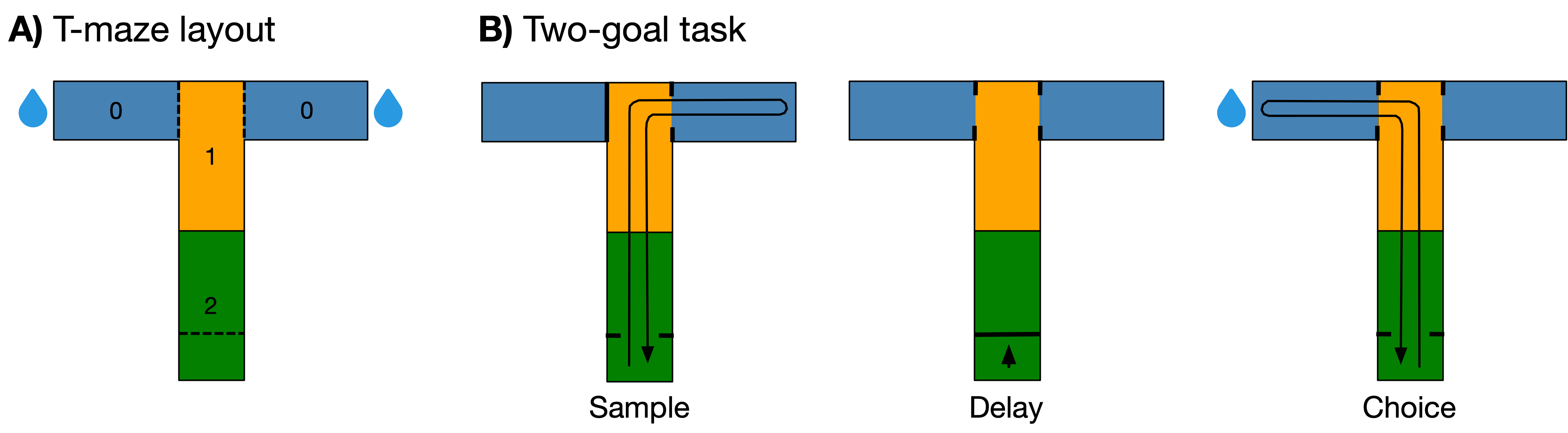}
\centering
\caption{ \textbf{A)} T-maze layout and discretization scheme. Maze locations are discretized into three sections (0/1/2). The holding zone is located at the base (2) while water spouts are placed in the arms (0). Dashed lines represent doors that can
%be used to 
constrain mouse movement. Location labels are symmetric between left and right arms.
\textbf{B)} The two-goal task consists of a sample phase, where the mouse must navigate from the holding zone to the open arm, a delay phase, where the mouse remains in the holding zone, and a choice phase, where the mouse must navigate to the opposite arm in an unrestricted maze.
}
\label{fig:t-maze}
\end{figure*}

Recording sessions took place in a T-shaped maze, illustrated in Figure~\ref{fig:t-maze}. The maze consists of a holding zone at the base, water spouts in each arm, and doors that can constrain mouse movements between the different sections. The mouse is first trained to navigate the maze in a specific sequence for a water reward upon successful completion. There are three phases in this two-goal T-maze task: sample, delay, and choice. In the sample phase, the mouse is placed in the holding zone and the door to one of the arms is opened, while the other door is left closed. The mouse must navigate from the holding zone to the end of the open arm and back to the holding zone. In the delay phase, a door is closed to keep the mouse in the holding zone for 10 seconds, while the doors to both arms are opened. In the choice phase, the mouse is released from the holding zone and must navigate to the end of the arm {\em opposite} from the one it entered earlier, where it receives a water reward. After the reward is delivered, the mouse then returns to the holding zone. To motivate completion of the task, water is restricted prior to task trials. In this task, mice must retain information about their past location to determine their immediate actions. Although the task is artificial, it does have some naturalistic components. For one, mouse movements are unconstrained, except during the delay phase, as the animals are allowed to roam freely throughout the maze and take arbitrary actions during the trials. Additionally, the duration of each phase, and hence the total trial duration, is uncontrolled, as events are triggered by mouse location.  All animal procedures followed the NIH Guidelines for Using Animals in Research and were approved by the NIH Animal Care and Use Committee.%, rather than predetermined.
% rather than by an experimentally predetermined timing. 

We began experiment sessions with a series of task trial recordings. Task trials were spaced out with a 30-second interval between the end of a trial and the start of the next trial. This was followed by a 5 minute waiting period before recording free running (FR) data. During FR, mice were allowed to freely explore the maze without being subject to any external constraints or stimuli, for as much time as needed to ensure that their trajectories covered the entire maze. Since data from FR and task were recorded in succession, we were able to track the same set of neurons between the two different contexts. Only mice that could successfully complete the task in 85$\%$ of trials, over three consecutive days, were 
%considered to be well-trained and 
used for data collection. Neural data and covariates were thus collected from 8 mice over 12 experiment sessions. We recorded between 38 and 104 (median 52) neurons per session. The data from one mouse were used to develop and validate our analysis approach, while the data from the remaining 7 mice were used solely to produce the results reported in Section~\ref{sec:experiments}.

To frame our data as a multivariate classification problem, we linearized mouse location so that both left and right arms positions were encoded symmetrically. Location labels were further discretized by dividing the maze into three sections (0/1/2), as shown in Figure~\ref{fig:t-maze}. Segments within the recordings were labeled as forwards (F) or backwards (B), depending on whether the mouse was moving towards the arms or the base of the maze. %respectively.
For our classification models, neural firings were used as the predictor, mouse location as the target, and movement direction as a confounding variable. Our method requires that the data have at least two independent subdatasets for each context. Task data were naturally split according to the start and end of each task trial to create several subdatasets, resulting in a median of 22.5 task subdatasets across sessions. As FR data have no natural split, we divided the data into two subdatasets by removing 20 seconds worth of time points from the middle of the time series. Full dataset properties can be found in Table \ref{tab:dataset_specs}.

\section{Theory}
\label{sec:theory}

\subsection{Testing for changes in neural code in a controlled experiment using decoding accuracies}
\label{sec:testing}

Suppose we have an experiment where we manipulate a \emph{label} -- a categorical variable such as the location of the animal -- in order to observe how the spike counts of a set of neurons $X$ depend on the label $Y$. We model the neural code by a decoding function $f \in \mathcal{F}$, which maps neural signals $X$ to possible label values $y$.  Throughout this paper, we take the family of decoding functions $\mathcal{F}$ to be the class of linear decoders, and we train them by fitting three different linear classifiers: a regularized Poisson decoder \cite{kim2003poisson, triplett2019probabilistic}, L2-regularized multinomial logistic regression \cite{pohar2004comparison, ng2002discriminative, maroco2011data}, and linear support vector machines (SVMs) \cite{chang2011libsvm}.  Our approach is agnostic to the function class and decoder, and could be applied to many other function families, such as K-nearest neighbors, random forests, neural networks, etc.

In order to test for differences in encoding between context A (e.g. task) and context B (e.g. free-running), we use the \emph{cross-classification accuracy} \cite{kaplan_multivariate_2015} (or simply \emph{cross-accuracy}) of decoding location by training a classifier in context A and comparing the performance of that classifier on independent test sets drawn from the same context (i.e., context A) versus a different context (i.e., context B).  Assume that the data consist of pairs $(X_i, Y_i)_{i=1}^n$ under context A and pairs $(X_i', Y_i')_{i=1}^n$ under context B,  where $Y_i$ and $Y_i'$ have been sampled uniformly. We split the data from both contexts into independent training and test sets to obtain estimates of accuracies $\text{Acc}^n$, defined as
\begin{equation}\label{eq:Acc}
\text{Acc}^n(X;Y) = \text{Pr}^n[f(X) = Y],
\end{equation}
and cross-accuracies $\text{XAcc}^n$, with cross-accuracy defined as
\begin{equation}\label{eq:XAcc}
\text{XAcc}^n((X;Y) \to (X';Y')) = \text{Pr}^n[f(X') = Y'].
\end{equation}
where $\Pr^n$ is the empirical probability on $n$ test data points and 
where decoder $f$ is trained on an independent training set  $(X_i^{train}, Y_i^{train})_{i=1}^m$. Ensuring set independence depends on the experimental setup; we give details for our setup in the supplement \S\ref{sec:partition}. In supplement \S\ref{sec:info_theory}, we provide the theoretical foundations for $\text{Acc}^n(X;Y)$ and $\text{XAcc}^n((X;Y) \to (X';Y'))$, and elaborate on how our constructions parallel classical analogues in information theory.

Given the definitions of in-context and cross-context accuracies (equations (\ref{eq:Acc}) and (\ref{eq:XAcc}), respectively), we then consider possible implementations of the cross-classification approach. A naïve approach would be to compare the accuracy of two classifiers, trained on training sets from two different contexts (e.g., one classifier trained on context A, and one on context B), and conclude that the contexts are significantly different if the performance of the classifiers differs on a third, independent set from one of the two context (e.g., a test set drawn from context A). However, this inference approach is erroneous, as random differences between the training sets used to train the classifiers will produce non-identical classifiers; as the test set grows to infinity, the difference in classifier performance will become significant, even if there is no difference in encoding. Alternatively, as detailed in \cite{kaplan_multivariate_2015}, one could test for differences in encoding in one direction (i.e., from context A to context B, or vice versa), by training a classifier in one context (e.g., context A), and using that trained classifier to test for differences in accuracy on an independent set in a different context (i.e., context B), as compared to test accuracy in the same context (i.e., context A). Here, we adapt this approach by testing differences in \textit{both} directions (i.e., from A $\rightarrow$ B and B $\rightarrow$ A) and averaging the result to create a single metric, called the symmetric decoding divergence (theoretical justification in \S\ref{sec:info_theory}). The estimate of this quantity is then
\begin{align*}
\text{SD}^n((X;Y), (X';Y'))=& \frac{1}{2}(\text{Acc}^n(X; Y) + \text{Acc}^n(X'; Y')
\\                          &- \text{XAcc}^n((X;Y) \to (X';Y'))
\\                          &- \text{XAcc}^n((X';Y') \to (X;Y))).
\end{align*}

The null hypothesis is that there is no difference in the encoding between the two contexts; that is, $X|Y=y$ has the same distribution as $X'|Y'=y$ for all $y \in \mathcal{Y}$. To test it, we need
%to know 
the distribution of $\text{SD}^n((X;Y), (X';Y'))$ under the null. %hypothesis.
Under mild conditions (see \S\ref{sec:Z.test.theory}), the central limit theorem implies that the four summands are approximately jointly multivariate normal.  Furthermore, under the null, all four accuracies are identically distributed, hence the mean is zero. That is, under the null, accuracy estimates from a classifier trained in the context A and tested on an independent set also drawn from context A (i.e., $\text{Acc}^n(X; Y)$) will have the same distribution as one trained in context A and tested on an independent set from context B (i.e., $\text{XAcc}^n((X;Y) \to (X';Y'))$). This implies that the $\text{Acc}^n(X; Y) - \text{XAcc}^n((X;Y) \to (X';Y'))$ term of $\text{SD}^n((X;Y), (X';Y'))$ has mean zero. The same applies in the reverse direction (i.e., to $\text{Acc}^n(X'; Y')$ and $\text{XAcc}^n((X';Y') \to (X;Y)))$), allowing us to control for random differences between classifiers trained in both contexts.

Combining these facts, $\text{SD}^n((X;Y), (X';Y'))$ is approximately normal with mean zero under the null.  This motivates the use of a one-sided Z-test on the mean of $\text{SD}^n((X;Y), (X';Y'))$.  The test is one-sided because $\mathbb{E}[\text{SD}^n((X;Y), (X';Y'))] \geq 0$ for most classifiers.
%
% [TODO: what is the issue with estimating the standard deviation?]
% CZ: It depends on the dependency between the data across different contexts which could be difficult 
%
For a given trained decoder and test set with $n_{test}$ examples, let $E_i$ be the binary indicator for a correct prediction on the $i$-th test example.
%of whether the prediction is correct for the $i$-th test example.
Assuming independence of the test examples, a consistent estimator of the standard deviation of the accuracy is $\frac{\hat{sd}(E_1,\hdots, E_{n_{test}})}{n_{test}}$. 
%(We will discuss how to deal with non-independence of test examples in [VIF section.])
We estimate the upper bound of the null distribution of $\text{SD}^n$ via the normal approximation to a binomial: 
% An upper bound for the standard deviation of the null distribution of $\text{SD}^n$ is
%
$\hat{\sigma}(SD^n) = \frac{1}{2}(\hat{\sigma}_1 + \hat{\sigma}_2 + \hat{\sigma}_3 + \hat{\sigma}_4)$
, where $\hat{\sigma}_1,\hdots,\hat{\sigma}_4$ are estimated standard deviations for the four test accuracies (proof in %supplement
\S\ref{sec:Z.test.theory}). While conservative as compared to more sophisticated approximation methods, this approach ensures type I error control and is adequately powered for most naturalistic neural recording tasks, as demonstrated in \S\ref{sec:experiments.testing}. The p-value for the one-sided Z-test is then $p = 1 - \Phi\left(\frac{SD^n}{\hat{\sigma}(SD^n)}\right)$, where $\Phi$ is the cumulative distribution function of a standard normal variate.
%
% \begin{figure*}[htb]
%   \centering
%     \includesvg[width=0.8\textwidth]{autocov_XM37_VIF11.svg}
%   \caption{In-context autocorrelation of the prediction error vector of a model fit on Task F data for mouse 37. The VIF is estimated as the first index (the indexing starts at 1) where autocorrelation becomes less than zero and occurs at index 11.}
%   \label{fig:autocov}
% \end{figure*}
%
% In the supplement \S\ref{sec:info_theory}, we elaborate on information axioms that $\text{Acc}^n$ satisfies, as well as how our constructions parallel classical analogues in information theory.

\subsection{Controlling for imbalances, temporal correlation, and confounds}

\paragraph{Label imbalances in a naturalistic setting}
\label{sec:control.matching}

Thus far, we have assumed the label distribution $Y$ is under control of the experimenter and sampled according to a fixed distribution. However, in our setting, the location $Y$ of the animal is not directly manipulated by the experimenter, so 
%that that 
the number of observations and the distribution of locations $Y$ within a recording will vary randomly. More importantly, these variations could be confounded with the context factor (task vs. free-running). Na\"{i}ve comparison of accuracies could produce false inferences, e.g. finding differences in accuracy due to the availability of more training data in one context, or the decoder defaulting to predicting the most frequent location in the data, rather than true differences in encoding.

\begin{figure*}[htb]
  \centering
    \includegraphics[width=\linewidth]{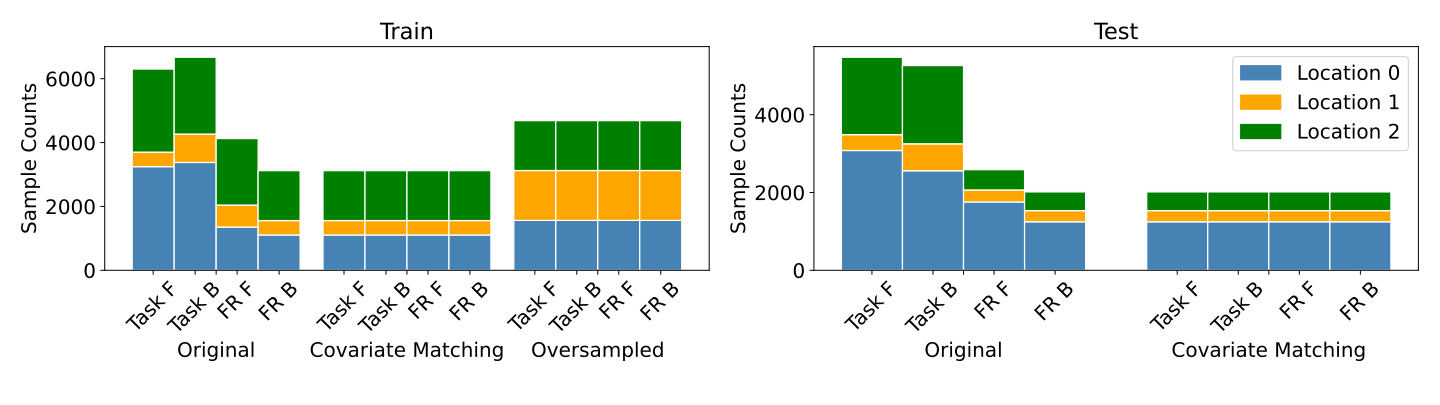}
  \caption{Illustration of covariate matching in the mouse data.  The goal of the method is to ensure that the label (Locations 0, 1, 2) has the same distribution for all decoders being compared (four decoders shown in Figure~\ref{fig:comparisons}, trained on Task and Free-Running, F(orward) and B(ackward)  respectively).  The left panel shows its application for training sets and the right for test sets. Each bar represents the raw data for one of the four decoders, and is divided into three sections to represent location label distribution within it. Both training and test sets are split by context and confound (movement direction), resulting in four splits.  The training partitions of the
  %preliminary
  datasets are subsampled to match the label counts across splits, and then oversampled to have the same number of examples per location. The testing partitions are likewise subsampled to match label counts across splits.}
  \label{fig:covariate.matching}
\end{figure*}

Therefore, we rely on a subsampling approach to create datasets that have matched sample sizes and label distributions, borrowing a method from causal inference called covariate matching (CM, \cite{rubin2006matched}) .
Since $Y$ is a confounding variable, applying CM results in an algorithm that matches test examples with the same label, between context A (task) and context B (free-running).
%that have the same label.
More generally, when testing for differences in encoding between any number of decoders, CM matches label distributions between all of them.  Algorithm details and proofs are given in the supplemental \S\ref{sec:cm_details}, but the intuition can be grasped from Figure \ref{fig:covariate.matching}.
%,which shows label matching applied to four encodings being compared.
The core idea is to make sure that the label distribution is uniform within both the training and test splits for both contexts. 
%However, covariate matching procedures are slightly different between train and test splits, due to their different goals and constraints.
For the training data, we maximize model performance by first matching label distributions across the splits of the training data resulting from CM, by randomly subsampling timepoints without replacement, and then oversampling the matched distribution to make each class label have the same number of examples. For the test data, we randomly subsample timepoints to include only unique ones.
%in a way that include only unique ones, so that we can later apply account for temporal correlation when estimating the standard error for hypothesis testing.

\paragraph{Temporal correlation}
\label{sec:temporal.correlation}

Neuron firing rates may not be independent over time, due to latency or history-dependent effects, low-frequency noise, physiological or environmental processes, or other time-dependent confounds.  We can account for this
%these effects
by assuming that the signal exhibits dependence over time. 
%We do this in our inferences by inflating
Specifically, we inflate variance in estimated accuracies by a variance inflation factor (VIF), where VIF=1 in the i.i.d. case.  The easiest way to estimate and correct for the VIF is to rely on domain knowledge, as we do. Otherwise, we estimate the VIF under the assumption that the dependence decays beyond a certain time interval, $k$. Then, the autocorrelation of the time series has expected value zero at all lags greater or equal to $k-1$, and $\hat{k}$ is the estimated VIF.

\begin{figure}[htb]
  \centering
    \includegraphics[width=0.45\textwidth]{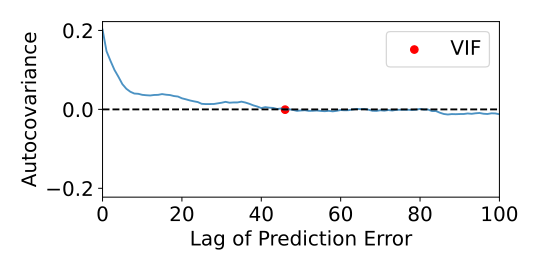}
  \caption{In-context autocorrelation of the prediction error vector of a model fit on Task F data, for mouse 37. The VIF is estimated as the first index
  %(indexing starts at 1)
  where autocorrelation is $\le0$
  %becomes less than zero
  (index 46).}
  \label{fig:autocov}
\end{figure}

We obtain the prediction error vector $E(t) = \textbf{I}\left(\hat{Y}(t) \neq Y(t)\right)$, where $t$ is the timepoint, and $\textbf{I}$ is the indicator function. We then compute the empirical autocorrelation as
$C_i = \frac{1}{n_{test} - i + 1} \sum_{t=1}^{n_{test}-i+1} (E(t) - \bar{E}) (E(t+i) - \bar{E})$, where $\bar{E}$ is the average error (one minus the empirical accuracy),
and we estimate VIF as the minimum $i$ such that $C_{i-1} \leq 0$ (see Figure \ref{fig:autocov}).
We note that since our motivation for estimating VIF is to correct for the effect of dependence on the inflated variance in the accuracy, our estimated VIF may not fully reflect the degree of temporal dependence in the data, but is only as conservative as necessary to ensure valid inference.
We provide theoretical analyses, simulation results, and pseudocode in supplement \S\ref{sec:VIF_details}.

\paragraph{Confounds}
\label{sec:confounds}
Often, there are other known factors that affect the encoding, besides the context factor that we wish to test.  In our example, it is possible that the direction the mouse was heading -- forward (F) vs. backward (B) -- might change the location encoding; therefore, to test for the effect of task (T) vs. free-running (FR), we need to control for the known effect of mouse movement direction.  False positives may result if the correlation between contexts and confounds spuriously boosts the detected effect size.  False negatives may also result, because failing to differentiate between multiple different encodings (a F encoding and a B encoding) can reduce the accuracy of classification.

Rather than train a single classifier per context, we stratify the data by confound -- split into forward and backward direction datasets -- and train a separate classifier within each stratum. 
%At this point, one could test a separate hypothesis for each stratum, but, as we illustrate in our data, it is also possible to perform a single combined hypothesis test of an overall context effect, averaging the effects within each context level.
%
Next, we perform an overall Z-test that averages the effects within each context level.  Let $c_1,\hdots, c_m$ be the various levels of the confound, and let $\delta_i, \sigma_i$ be the estimated symmetric decoding divergence and estimated standard deviation for the $i$th confound level, respectively. We can perform a one-sided Z-test by averaging the means and averaging the standard deviations, hence computing $p = 1 - \Phi\left(\frac{\sum \delta_i}{\sum \sigma_i}\right)$.  We give more detailed remarks and proofs in the supplement \S\ref{sec:Z.test.theory}.

\subsection{Alternative hypothesis tests}
In contrast to our approach, it is also possible to test for differences in encoding using classical univariate and multivariate two-sample hypothesis tests, albeit without the benefit of controlling for all the confounds that we address in our method. To do so, we first combine observations across subdatasets for a given subject and context, and then stratify the firings for each context by the label (e.g. location) and confound (e.g. movement direction). For each stratum, we apply the two-sample test, using the spike counts from each context as the samples, to compute a $p$-value for that stratum. If the minimal $p$-value across strata is less than or equal to the threshold given by the conservative Bonferroni correction for the number of strata tested, we reject the null hypothesis that the encoding is the same for both contexts.  Otherwise, we accept the null hypothesis.
%after correcting for multiple comparisons using the conservative Bonferroni correction.
The univariate tests we consider are the unpaired $t$-test (``$t^2$''), Kolmogorov–Smirnov (``$\text{KS}$''), and chi-squared (``$\chi^2$''). The multivariate tests are Hotelling's $T^2$ (``$T^2$''), as well as two independence tests adapted for use as nonparametric two-sample tests of distributional equivalence \citep{panda2019nonparametric}
%, as described in \cite{panda2019nonparametric}
: distance correlation (``$\text{DCorr}$''), and mean maximum discrepancy (``$\text{MMD}$''). We provide details about tests and implementations \S\ref{sec:alternative.hypothesis.tests}.

% {\color{white} [extra text]}

\subsection{Recovery of neural tuning curves with a Bayesian Poisson decoder}
% shorter, but cannot pull 3.2 up!
%The Poisson decoder assumes neuron spike counts are a Poisson process with a stimulus-dependent intensity parameter, with each neuron conditionally independent given the stimulus.
% keep the sentence if we can't pull 3.2 up into page 4
We use the Bayesian Poisson decoder as our primary classifier of interest, because in addition to allowing us to test for differences in encoding between contexts, its parameters can be interpreted as discretized single-neuron tuning curves. The decoder is based on a model where neuron spiking activity is a Poisson process with a stimulus-dependent intensity parameter, with each neuron conditionally independent given the stimulus. That is,
$X_t|Y_t=y \sim \text{Poisson}(\lambda_y)$,
where $Y_t$ is the location at time $t$, $X_t$ is the spike count for the neuron, and $\lambda_y$ is the location-dependent intensity for the neuron. As a function of $y$, this is a \emph{discretized tuning curve} that shows, for each location, how much the neuron fires while the mouse is located there.  This model is also known as \emph{Poisson naïve Bayes} \cite{kim2003poisson}, which is a special case of the linear-nonlinear-Poisson model \cite{triplett2019probabilistic} in the special case of a single categorical regressor.
%
%allows one to test for differences in encoding, and also to conduct post-hoc analyses at the individual neuron level.  This is because its parameters can be interpreted as discretized single-neuron tuning curves.  The decoder 
%
The estimation of $\lambda_y$ is done using a Bayesian conjugate Gamma distributed prior with parameters $\lambda_0$, the prior intensity, and $n_0$, the prior sample size \cite{gelman2013bayesian}.  Given data with $p$ neurons, $(X_t^1,\hdots,X_t^p, Y_t)_{t=1}^T$, the parameter $\lambda_{i,j}$ is estimated as
\[
\lambda_{i,j} = \frac{\lambda_0 n_0 + \sum_{t=1}^T I(Y_t = j)X_t^i}{n_0 + \sum_{t=1}^T I(Y_t = j)}.
\]
Conversely, the decoder works by computing posterior probabilities of $Y$ given the estimated Poisson distributions.  Hence, for a new observation $X_*$, we predict the most likely label (location) $Y_*$ as
\[
Y_* = \text{argmax}_j \sum_{i=1}^p -\log(X^i_*!) -\lambda_{i,j} +  X^i_*\lambda_{i,j}.
\]
Thus, the Poisson decoder is a linear classifier because the label is determined by comparing linear functions of the input.
To choose the hyperparameters $n_0$ and $\lambda_0$, we apply cross-validation on the training set over a grid of candidates, and pick the pair yielding the highest cross-validated accuracy.

% in-column plot
% \begin{figure}[!htb]
%   \centering
%   \includesvg[scale=0.45]{TuningCurves2.svg}
%   \caption{The simulated mouse moves continuously forwards from location 0 to 2 and backwards from 2 to 0, in two contexts (task vs. free-running).
%   {\bf a:} Ground truth 1D location tuning curves for different contexts (neurons: sensitive to location regardless of direction, sensitive only when moving towards goal, insensitive). {\bf b:} Estimated tuning curves from model operating on discretized locations, with Poisson decoders fit to each combination of context and direction.}
%   \label{fig:tuning.curves}
% \end{figure}

\section{Experiments and Results}
\label{sec:experiments}

\subsection{Synthetic Data Experiments}
\label{sec:experiments.synthetic}

%We demonstrate our method on two different synthetic data-generating setups. The simple synthetic data model allows us to demonstrate the principles of our method as transparently as possible, while the more complex data model allows us to stress-test our approach under realistic data settings.

%\emph{Simple experiment}. 

\begin{figure*}[!htbp]
  \centering
  \includegraphics[width=\linewidth]{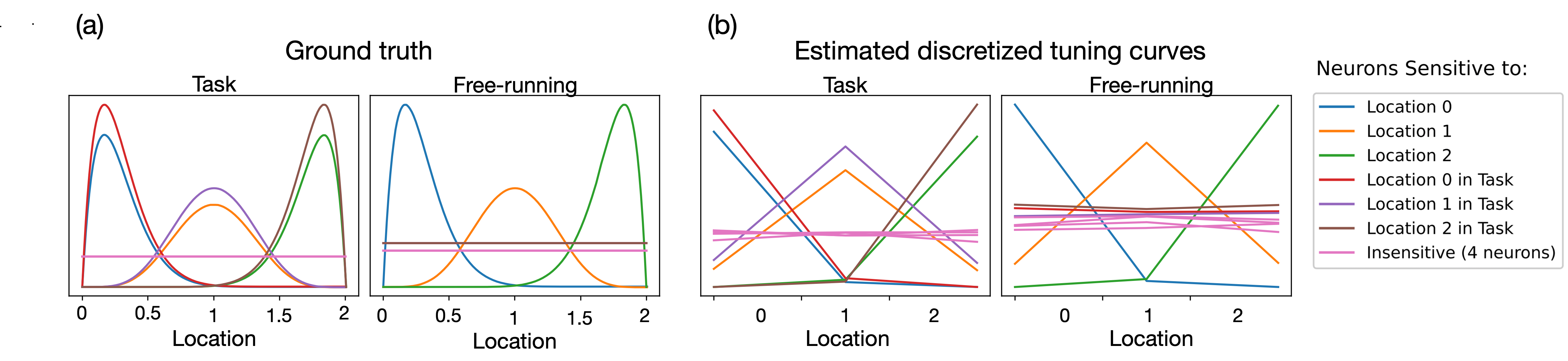}
  \caption{The simulated mouse moves continuously forward from location 0 to 2 in two contexts (task and free-running).
  %independent Poisson neurons with the displayed tuning curves, with 
  {\bf a:} Ground truth 1D location tuning curves for different contexts. {\bf b:} Estimated tuning curves from model operating on discretized locations, with Poisson decoders fit to each combination of context and direction.}
  %The plots on the right show the ground truth tuning curves used to generate simulated data. The plots on the left show the estimated discretized tuning curves obtained from the Poisson decoder.  The simulation consisted of independent Poisson neurons with the displayed tuning curves, given a simulated mouse that went continuously from location 0 to 2 and back to 0 at a uniform speed over the course of 122 time steps. The location was discretized to three equally spaced intervals and Poisson 
  \label{fig:tuning.curves}
\end{figure*}

\paragraph{Simulation 1: Recovery of neuron tuning curves} To validate the recovery of the location tuning curve of each neuron from the Poisson decoder weights, we created a dataset simulating a mouse moving continuously forward from location 0 to 2 in two contexts (task and free-running). We generated 600 data points with the mouse moving at a constant speed. The mouse had 10 neurons, 4 of which were insensitive to location, and 6 which were sensitive to location. The tuning curves are shown in Figure~\ref{fig:tuning.curves}a. Ground truth neuron location tuning curves were modeled with beta distributions: $\alpha$ and $\beta$ shape parameters control location specificity, and a tuning curve scale parameter controls the intensity of neuron firing. We drew neuron spike counts $X_t$ from their Poisson random variables as the mouse moved. For neurons that are sensitive to a particular context, the Poisson firing rate is specified by the probability density function of the ground truth tuning curve. Otherwise, the firing rate is uniform across all locations. A Poisson decoder trained on data discretized to three regions successfully recovered corresponding discretized tuning curves (see Figure~\ref{fig:tuning.curves}b). 
 
% column span plot

\begin{figure}[!htb]
    \centering
    \includegraphics[width=\linewidth]{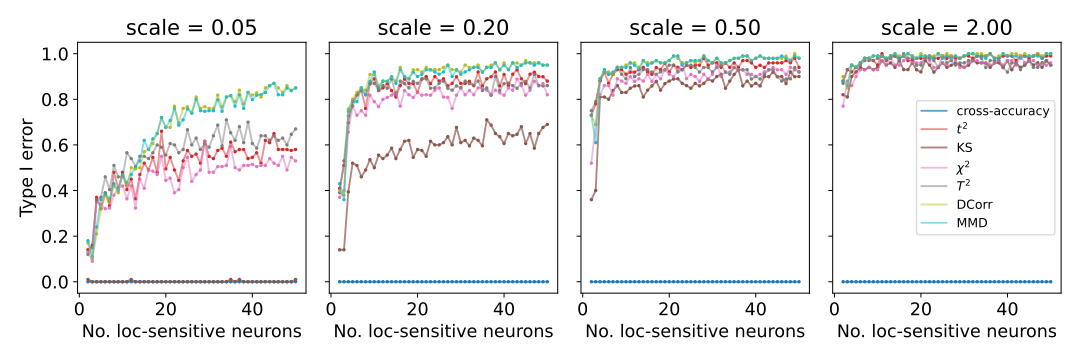}
    \caption{Type I error of our cross-accuracy-based approach and alternative hypothesis tests by the number of location-sensitive, non-context-dependent neurons and scale factor, evaluated across the grid of neuron numbers $\{ 2, 3, ..., 50 \}$ and tuning curve scales $\{ 0.05, 0.20, 0.50, 2.00 \}$. We consolidate the cross-accuracy-based tests (Poisson decoder, logistic regression, and SVM), using both a fixed VIF of 12 and an estimated VIF, into a single line ("cross-accuracy") because all had a type I error rate of 0 across parameters. }
    \label{fig:sim.type1.error}
\end{figure}

\paragraph{Simulation 2: Type I error rate} To evaluate whether our approach is sufficiently conservative and robust to the temporal dependence present in our experiment, we compared the type I error of our approach with the alternative hypothesis tests on simulation data. To do so, we simulated sets of neurons that were responsive to location, but had the same tuning curves in task and free-running trials. Because the neurons' tuning curves were the same in both contexts, any result that found a significant difference in encoding between contexts was a false-positive. We estimated the type I error rate, as a function of the number of location-sensitive neurons, by running simulations with 100 random seeds for each parameter setting, and recording the proportion of $p$-values for which $p \leq \alpha$, where $\alpha=0.05$ is the significance level.
%($0.05$ in this analysis).
We also considered a variety of tuning curve scales (see \S\ref{sec:synthetic.experiments}), such that higher scales corresponded to higher firing rates. The tuning curve scales selected represent a cross-section similar to those observed in the real data (see Figure \ref{fig:experimental.firing.rates}). As shown in Figure \ref{fig:sim.type1.error}, the alternative hypothesis tests suffered from inflated type I error rates, particularly as the number of context-dependent neurons and tuning curve scale increased. In supplement \S\ref{sec:synthetic.experiments} we examine how the sensitivity of the cross-accuracy test varies depending on the classifier used.

\subsection{Experiments on Neural Data}
\label{sec:experiments.formation}

% [FP: agree]
%[CZ: I think the first two sentences are unnecessary.]  Context effects are isolated by a careful data splitting procedure. Preprocessed T-maze data is first divided between the two contexts, and then further stratified along confounding variables. [CZ: Begin here.] 
\paragraph{Dataset formation}
As location encoding may depend on the direction of mouse movement, we considered  direction a confound, and stratified the data by
%according to 
forwards (F) and backwards (B) movement. This yielded four splits -- Task F, Task B, FR F, and FR B -- in each mouse. For each split, we formed independent train and test sets of approximately equal sizes (see \S\ref{sec:partition}); these were then subsampled with the covariate matching procedure in 
%using algorithm \ref{alg:train_test_partition}.  
Section~\ref{sec:control.matching},
%in algorithm \ref{alg:train_test_split}
to address confounds such as differences in dataset sizes and label distributions.

%eliminate confounding dataset properties such as differences between dataset sizes and label distributions. 

% Text When table was available
% To ensure dataset independence between train and test splits, each preliminary dataset is bifurcated into two partitions, one reserved for training and the other for testing, using algorithm \ref{alg:train_test_partition}, illustrated in Figure~\ref{fig:covariate.matching}A (left) and B (left). In order to ensure sufficient training and test data, the two partitions are of approximately equal sizes. Partitions are then subsampled with the covariate matching procedure in algorithm \ref{alg:train_test_split} to eliminate confounding dataset properties such as differences between dataset sizes and label distributions, as illustrated in Figure~\ref{fig:covariate.matching}A (middle). Covariate matching procedures are different between train and test splits due to

%% Moved to Appendix under Covariate Matching
% 

%\subsection{Multivariate Classification}
%\label{sec:experiments.classification}

\begin{figure*}[htb]
%[htb]
%
  \begin{subfigure}{0.45\textwidth}
  
  \centering
   \includegraphics[width=0.8\linewidth]{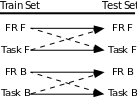}
    \caption{ Decoding Hypotheses}

    \label{fig:comparisons}
  
  \end{subfigure}%
  \hspace*{\fill}   % maximize separation between the subfigures
  \begin{subfigure}{0.55\textwidth}
    \includegraphics[width=\linewidth]{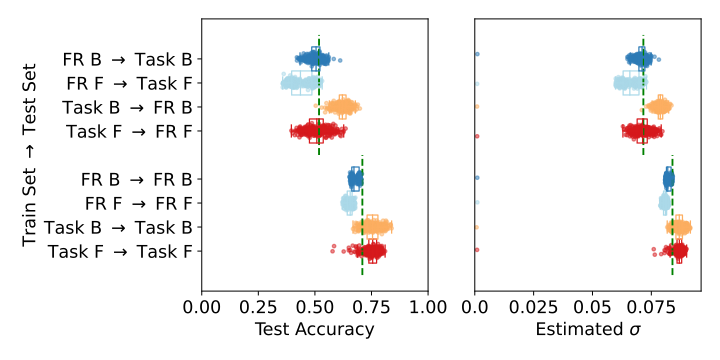}
    \caption{Aggregating Hypotheses} 
    \label{fig:accuracy.distribution}
  \end{subfigure}%
% \caption*{\textbf{Figure \ref{fig:accuracy.distribution}} } \label{fig:comparisons_and_boxplots}
\caption{ {\bf a: Decoding hypotheses to determine the presence of context effects in location encoding.} Solid lines represent accuracies of classifiers trained and tested in the same context, dashed lines represent accuracies when they are trained and tested in different contexts. Decoder statistics are pooled together for a combined hypothesis test. {\bf b: Test accuracy and standard error distribution for each prediction direction in Figure \ref{fig:comparisons}, over 400 random seeds.} Top/bottom four prediction directions represent cross-context and same-context measurements, respectively. Boxplots of the same color group together corresponding same-context and cross-context decoder statistics. Dashed green lines show the average value across decoders and seeds.}
\end{figure*}

\paragraph{Classifier training}
We trained a Poisson decoder
%Poisson  Na\"{i}ve Bayes classifiers 
to predict discretized maze locations from spike counts of the recorded PFC neurons in each mouse. To account for stochasticity of neural activity (e.g. informative neurons might not fire at the precise moment a mouse is in a location), we aggregated a small amount of past information when generating a prediction at a given moment. %The features for each classifier were the neuron average firing rates in all ten 40ms bins during a short interval of 400ms, preceding and up to the moment for which a prediction was generated.
% correct but let's save some space
%Time lag should ideally be a short interval on the time scale of the behavior so that neural information from other locations in the past is not leaked into the present prediction.
In all our classifiers, the window was set to 10 timepoint samples (roughly 0.4 seconds), based on prior experiments with data from animals not used in this paper, and hyperparameters were fit using 5-fold cross validation. The input features within the interval with time lag $l$ are denoted as 
$\widetilde{X}(t, l) = \left[ X(t), \dots, X(t-l) \right]$, 
where $X(t)$ contains the spike counts for all neurons at timepoint $t$. 
%Given $\widetilde{X}(t, l)$, 
The model was trained to predict the location label $Y(t) \in \{0, 1, 2\}$ as in Figure \ref{fig:t-maze}. We specified a uniform prior with two parameters, $\lambda_n \in \{0, 1, 5, \ldots, 100, 500, 1000\}$ for the prior number of samples and $\lambda_{\text{rate}} \in \{\frac{0}{2}, \frac{1}{2}, \dots, \frac{19}{2}, \frac{20}{2}\}$ for the prior firing rate, to prevent overfitting. We also considered two other classifiers: a logistic regression model and a linear support vector machine (SVM), implemented using Scikit-learn \cite{scikit-learn}. For both the logistic regression and SVM classifiers, the hyperparameter was the inverse regularization strength, selected from the grid $\{ 10^{-4}, 10^{-3}, ..., 10^4 \}$.
%$\lambda_n \in \{0, 1, 5, 10, 50, 100, 500, 1000\}$ and $\lambda_{\text{rate}} \in \{\frac{0}{2}, \frac{1}{2}, \dots, \frac{19}{2}, \frac{20}{2}\}$.

% We use L2 regularization to prevent overfitting, optimizing $\lambda$ for each train split separately, using a 5-fold cross validation grid search with $\lambda \in \{2^{-20}, 2^{-19}, \dots, 2^{19}, 2^{20}\}$. 

\paragraph{Hypothesis Testing}
\label{sec:hypothesis.testing}
To test if location encoding is affected by context, we considered the symmetric decoding divergences of four classifiers trained in each of the Task F, FR F, Task B, and FR B training sets, as shown in Figure~\ref{fig:comparisons}, and performed an \emph{overall Z-test} (\S\ref{sec:control.matching}) across all decoders.
%in Table~\ref{fig:comparisons}.
To account for temporal correlation, we set $\text{VIF}=12$, determined by domain knowledge (maximum overlap
%number of overlapping time points
in the lag windows of a decoder operating in adjacent test examples, plus three, %corresponding to
a duration of 0.5 seconds). To reduce the variance due to resampling, we averaged results from 400 repetitions of the procedure with different random seeds (see \S\ref{sec:Z.test.theory}), each yielding a training and test set.

\subsection{Results}
\label{sec:experiments.testing}

We applied our procedure to data from 10 recording sessions of seven held-out test mice, in addition to data from two sessions of the one mouse used in development. The results are shown in Table~\ref{tab:results}. For each session, the table shows the number of neurons recorded (different neurons per session, in each animal). The next two columns -- $\obar{\text{acc}}_{\text{same}}$ and $\obar{\text{acc}}_{\text{cross}}$ -- show the average performance of the four decoders in Table~\ref{fig:comparisons}, across 400 random seeds, and the median of a conservative VIF estimate. The following columns show the $p$-value for our test statistic using a Poisson decoder with $\text{VIF}=12$, and the effects on the $p$-value of using estimated VIF, not using covariate matching, and not using confound stratification. In Table \ref{tab:results.full} we include the logistic regression and linear SVM results as well. These results corroborate those of our simulation analysis in \S\ref{sec:synthetic.experiments}, which finds that these classifiers have lower power than the Poisson decoder.

\begin{table*}[ht!]
\centering
\caption{Results of the testing procedure to detect context effects in location encoding, for each of two development and 10 held-out mouse sessions, with a different set of neurons per session. The values $\obar{\text{acc}}_{\text{same}}$ and $\obar{\text{acc}}_{\text{cross}}$ (the estimates of $\text{Acc}_{same}$ and $\text{Acc}_{cross}$) are the average performance of the four decoders in Figure~\ref{fig:comparisons}, across 400 random seeds. The following column, "$\text{est. VIF}$", contains the VIF estimated using the Poisson decoder. Subsequent columns show the the $p$-value for our test statistic using $\text{VIF}=12$, a VIF estimated from data, and the individual effects of not using VIF, covariate matching, and confound stratification. }

\resizebox{\textwidth}{!}{\begin{tabular}{@{\extracolsep{4pt}}lcccccccc@{}}
\\
\toprule
   & & & & \multicolumn{5}{c}{Poisson decoder $p$-values} \\
  \cmidrule{5-9}
  \thead{mouse/sess.\\(\# neurons)}  & \thead{$\obar{\text{acc}}_{\text{same}}$\\ ($\bar{\sigma}_\text{same}$)} & \thead{$\obar{\text{acc}}_{\text{cross}}$\\ ($\bar{\sigma}_\text{cross}$)} &  \thead{est.\\VIF} & \thead{fixed\\VIF\\=12} & \thead{using\\est.\\VIF} & \thead{without\\VIF} & \thead{without\\covariate\\matching} & \thead{without\\conf.\\strat.} \\
\cmidrule{1-9}
   
37/1 (41)	& 0.75 (0.007)	& 0.60 (0.007)	& 48	& \textbf{1.72e-3} & 0.081	& 1.92e-24 & 5.52e-4 & 1.06e-04 \\
37/2 (38)	& 0.76 (0.007)	& 0.64 (0.007)	& 42	& \textbf{5.41e-3} & 0.135	& 5.29e-19 & 6.10e-3 & 2.55e-04  \\ \\

36/1 (72)	& 0.81 (0.007)	& 0.66 (0.009)	& 40	& \textbf{3.53e-3} & 0.091	& 5.20e-21 & 1.07e-3 & 1.27e-04  \\
36/2 (96)	& 0.71 (0.009)	& 0.52 (0.009)	& 58	& \textbf{1.43e-3} & 0.126	& 2.59e-25 & 5.84e-4 & 1.88e-14  \\
38/1 (52)	& 0.61 (0.009)	& 0.35 (0.009)	& 57	& \textbf{4.71e-5} & 0.042	& 5.35e-42 & 9.25e-7 & 2.33e-12 \\
38/2 (51)	& 0.56 (0.009)	& 0.38 (0.009)	& 45	& \textbf{2.72e-3} & 0.158	& 2.97e-22 & 4.58e-4 & 4.95e-09  \\
39/1 (42)	& 0.63 (0.008)	& 0.48 (0.008)	& 53	& \textbf{3.15e-3} & 0.112	& 1.51e-21 & 5.19e-3 & 1.23e-03  \\
40/1 (104)  & 0.75 (0.007)   & 0.65 (0.008)   & 56	& \textbf{3.63e-2} & 0.248	& 2.47e-10 & 7.94e-2 & 1.02e-09  \\
43/1 (45)   & 0.84 (0.007)   & 0.78 (0.007)   & 39	& 1.13e-1          & 0.269	& 1.40e-05 & 1.04e-1 & 1.69e-04 \\
44/1 (42)	& 0.70 (0.007)   & 0.61 (0.008)   & 50	& \textbf{4.78e-2} & 0.213	& 3.88e-09 & 2.88e-2 & 3.15e-03  \\
46/1 (56)	& 0.81 (0.007)   & 0.73 (0.008)   & 49	& \textbf{4.41e-2} & 0.248	& 1.74e-09 & 1.38e-2 & 1.44e-03 \\
46/2 (57)	& 0.76 (0.006)   & 0.63 (0.007)   & 56	& \textbf{1.61e-3} & 0.102	& 9.66e-25 & 2.97e-4 & 2.97e-10  \\

  \bottomrule
\end{tabular}}
\label{tab:results}

\end{table*}

Out of 10 test sessions, we could reject the null hypothesis of no difference in location encoding in nine of them (significance  $p=0.05$). For the remaining test session ($43/1$), the accuracy within the same context was always higher than across context, but with a smaller gap than in the others. 
% Note that, if we used the conservatively estimated VIF, most would have remained significant. 
Finally, 
%note that, 
without incorporating VIF correction or confound stratification, the $p$-values became much smaller, suggesting the addition of spurious contrast due to these confounds. 
%Without confound stratification, it would have been hard to see {\em any} difference between those conditions.

\section{Discussion}
\label{sec:discussion}
In this paper, we introduced a systematic approach for using classifiers to detect differences in neural encoding of information across experimental contexts. 
We combine a variety of methods -- covariate matching, variance correction factor (VIF) estimation, splitting and combining by confound, combining over random seeds -- which can be incorporated as needed into a pipeline for testing context effects for a given dataset. If the stimuli are presented in a controlled order but time series correlations exist, then we need VIF estimation but not label matching.  
On the other hand, if the distribution of stimuli is random but they are spaced out far enough in time for autocorrelations to be a non-issue, then label matching is needed but not VIF estimation.  In simulation experiments, we showed the necessity of combining these confound corrections. All the alternative testing approaches we compared against, both univariate and multivariate, suffered from inflated type-I error due to the violation of the assumption that data points are independent and identically distributed (i.i.d.). Moreover, the univariate $t^2$ test suffered from violation of its gaussianity assumption for the input data, and hence had even higher type I error rate than the KS test. While there are potential remedies to correct for these issues in classical hypothesis tests, for example via estimating effective degrees of freedom \cite{efron2012large} or bootstrapping \cite{wilks1997resampling}, this requires nontrivial modifications to the existing methods and is beyond the scope of this work. Our approach, by contrast, maintained nominal type I error control in all simulation settings,  with all three classifiers considered. Additionally, while prior work has shown classification-based methods may be underpowered relative to classical and modern two-sample tests, particularly in cases with a relatively small sample size \cite{rosenblatt2021better}, we mitigate this effect by using resampling in our dataset formation.

% We note also that our method for inferring classifier generalization accuracy from performance on correlated time series data may be interesting to the wider machine learning literature. 

%It is possible that many of these techniques (label matching, VIF estimation) could be applied to testing approaches not based on classifier accuracies (e.g. using models of receptive fields for individual neurons, which we did not have in this instance).
%
Meanwhile, in real data, we demonstrated the practical applicability of this approach to the problem of testing whether location encoding in the mouse PFC differs between task and free-running contexts. We found significant effects ($p\leq 0.05$) in seven out of eight mice and 11 out of 12 sessions using our assumed VIF based on prior knowledge. In contrast, using the VIF estimated from data gave significant results in only one session, suggesting that the estimated VIF may be too conservative.  By only counting independent data points, we ignore the information contributed by data with small but non-zero correlations, and hence overestimate the VIF.  The other extreme of assuming independence seems unwarranted, giving extremely low $p$-values for all sessions.  Assuming a VIF based on prior knowledge gave more reasonable results, where $p$-value negatively covaried with empirical difference in accuracies.
%Large p-values in three sessions tend to correspond to low within-context prediction accuracy. As accuracies vary greatly even within the same animal, these variations in are plausibly due to variation in neuron sampling: sometimes the probe is placed near one or more clusters of neurons that are especially sensitive to location and sometimes not.
The effect of covariate matching was to increase $p$-values in some sessions, while decreasing $p$-values in other sessions. This indicates that there is between-subject variation in the effect of label imbalance on transfer accuracy.  
% p-values were also extremely low, as large differences in dataset size and label distribution contributed to worse transfer accuracies, which misleadingly widen the gap between $\obar{\text{acc}}_\text{same}$ and $\obar{\text{acc}}_\text{cross}$. 
We see evidence of the necessity of stratifying by confound (forward vs. backwards) in the fact that failing to stratify lead to lower $p$-values in all 12 subjects, suggesting the existence of correlations between movement direction and context, which would invalidate the inference if uncontrolled.
% without it, we see inconsistent p-values (likely due to the effect of training and testing on mixture encodings.)

Although the primary motivation of our method is to detect context effects, we show that using a Poisson decoder
%Naive Bayes classifier as the decoder function 
gives us the additional ability to recover the discretized tuning curves for each neuron. This can be used to further investigate the behaviors of the neuron population of interest in different contexts. Some limitations of our approach include a lack of explicitly modelling neuron spiking history effects \cite{pillow_model-based_2011} and cross-neuron interactions. Hence, our method could be underpowered for data where such dynamics play a role. In future work, it may be possible to  address this limitation by extending the covariate matching approach to match for both label and history.
%
% make mcnemar's discussion more concrete. 
% YC: expanded to include what we discussed this morning. 
%An alternative statistical test that we considered was Mcnemar's test for differences in model performance. One obstacle that we faced was determining a method that allows us to combine different hypotheses into a single statistic. With the Z-test, combining accuracies is straightforward since the test accuracy is a property of the entire decoder. With Mcnemar's test, we only consider the differences in mistakes between paired decoders and it is unclear how to combine these paired tests across several hypotheses. However, our basic approach could potentially be improved by substituting the Z-test for McNemar's test given that we can develop a new theory for how to combine tests.
%
The ability to test for a context effect segues neatly into testing individual neurons or groups of neurons to see which specific subset of the neurons is affected by the context, which we are pursuing. Additionally, we use an extremely general bound in estimating the variance of our test statistic; while this approach requires minimal assumptions, and we demonstrate it is sufficient for our dataset, power could be improved by using a more sophisticated approximation method. Beyond this, we could develop a parametric model for contexts, and explicitly describe the effect of context on the encoding as a function of the context features. This would allow prediction of the effect of a new context on encoding.

% {\color{white} [extra text]}

% [FP: may copy from here]
%\paragraph{Pipeline overview}
%The methods that we presented (label matching, VIF estimation, splitting and combining by confound, combining over random seeds) can incorporated as needed into a pipeline for testing context effects for a given dataset.  If the stimuli are presented in a controlled order but time series correlations exist, then we need VIF estimation but not label matching.  On the other hand, if the distribution of stimuli is random but they are spaced out far enough in time for autocorrelations to be a non-issue, then label matching is needed but not VIF estimation.  In our data, we have an uncontrolled target, time series correlations, and an additional confound, so in the following section we present a example of a complete analysis incorportating all of the tools that we developed.

\section*{Acknowledgements}
This work was supported by the National Institute of Mental Health Intramural Research Program (ZIC-MH002968,ZIA-MH002882).  All animal procedures followed the US National Institutes of Health Guidelines Using Animals in Intramural Research and were approved by the National Institute of Mental Health Animal Care and Use Committee. This work utilized the computational resources of the NIH HPC Biowulf cluster (http://hpc.nih.gov).  We thank our anonymous reviewers for their helpful comments.
%, and especially reviewer 8ips for suggesting a useful reference.

%\section*{References}

\bibliographystyle{unsrt}
\bibliography{decoding}

%%%%%%%%%%%%%%%%%%%%%%%%%%%%%%%%%%%%%%%%%%%%%%%%%%%%%%%%%%%%
\newpage
\appendix
\section{Appendix}
\label{sec:appendix}

\localtableofcontents

\subsection{Remarks on Terminology}\label{sec:terminology}

\paragraph{Defining encoding}
Here we have defined encoding as any set of conditional distributions of $X|Y=y$.  However, from a neuroscience perspective one might want a stricter condition for $X$ to encode $Y$ beyond the existence of conditional distributions.  For instance, one might require that the conditional distribution of $X$ actually depend on $Y$ non-trivially, meaning that $X$ actually contains information about $Y$. Even stronger, one might require conditional distributions of $X$ to be able to separate distinct values of $Y$.  And one might also expect that $X$ exclusively encodes for $Y$, and no other target variable of interest, in the sense that $Y$ causally screens off $X$ from other potential target variables.\footnote{To be precise, supposing we have potential targets $Y^{(1)},\hdots,Y^{(m)}$, then $X$ exclusively encodes for $Y^{(j)}$ if $\{Y^{(j)}\}$ is a Markov blanket (Definition \ref{def:MB}) for $X$ with respect to $\{X, Y^{(1)},\hdots,Y^{(m)}\}$.} However, given the lack of a standardized definition of encoding, we have adopted a very general definition.

\paragraph{Statistical control}
Our work provides \emph{statistically principled control against false positives} in the following sense.  We adopt a criteria for evaluating false positive rate, which is the \emph{type I error at level $\alpha$}, i.e., the probability that the $p$-value produced by our test is less than or equal to $\alpha$ under a \emph{null hypothesis} where there is no difference between encodings.  Equivalently, we control type I error at all levels $\alpha$ if the $p$-value under a null distribution stochastically dominates a uniform distribution; that is,
\[
\Pr[p \leq x] \leq \Pr[U \leq x]
\]
for all $x \in [0,1],$ where $U$ is a Uniform([0,1]) random variate.

We assess the type I error control of our methods in two ways.  (1) For a specific idealized situation with no confounds and asymptotically increasing sample size, we prove that the $p$-value converges in distribution to a uniform distribution (by showing that the test statistic converges to a normal distribution with known mean and variance.)  (2) For more general situations with confounds, we have present theoretical arguments for why our heuristically motivated techniques should succeed in maintaining conservative (i.e., type I error rate less than or equal to $\alpha$) inference under the appropriate conditions.  However, we have not proved type I error control for the general case.  Rather, we have verified that the procedure is conservative using simulation studies that are described in \ref{sec:synthetic.experiments}.

\subsection{Information theory}\label{sec:info_theory}

Suppose we have an experiment where we manipulate a target (e.g. the location of the animal) in order to observe how the firing rates of a set of neurons $X$ depend on the target $Y$. As we describe in \S\ref{sec:testing}, we quantify the information in the encoding using the optimized $\mathcal{F}$-decoding accuracy
\[
\text{Acc}^*_{\mathcal{F}}(X;Y) = \sup_{f \in \mathcal{F}} \Pr[f(X) = Y].
\]
where $f: \mathcal{X} \to \mathcal{Y}$ are functions within a function class $\mathcal{F}$.

The optimized $\mathcal{F}$-decoding accuracy satisfies some important properties of information:
\begin{itemize}
    \item[I1.] It is non-negative.
    \item[I2.] Fixing the distribution of $Y$, it takes a minimal value when $X$ and $Y$ are independent.
    \item[I3.] $\text{Acc}^*$ takes a maximal value 1 when $Y$ is a deterministic function of $X$, $Y=f(X)$, and $f \in \mathcal{F}$.
    \item[I3.] Fixing the distribution of $Y$, $\text{Acc}^*(\cdot;Y)$ satisfies \emph{monotonicity}: the information value of a set of a neurons increases as more neurons are added.
\end{itemize}

We state and prove these properties in the following theorem.

\begin{theorem}\label{the:info_axioms}
Let $Y$ be a distribution on a discrete set $\mathcal{Y} = \{1,\hdots, k\}$. Assume that $\mathcal{G}$ is a function space for maps $g: \mathbb{R}^{2p} \to \mathcal{Y}$.  Further assume that $\mathcal{F}$ is a function space for maps $f: \mathbb{R}^p \to \mathcal{Y}$ such that for all $f \in \mathcal{F}$, there exists $g \in \mathcal{G}$ such that $f(x) = g(\tilde{x})$ for all $x \in \mathbb{R}^p$ and all $\tilde{x} \in \mathbb{R}^{2p}$ such that $\tilde{x}_i = x_i$ for $i=1,\hdots, p$.  Define $\text{Acc}^*_{\mathcal{F}}(X;Y) $ as in \eqref{eq:Acc}.   Then the following hold.

(i) $\text{Acc}^*_{\mathcal{F}}(X;Y) \geq 0$.

(ii) If $X' \perp Y$, then
\[
\text{Acc}^*_{\mathcal{F}}(X';Y) = \max_{y=1}^k \Pr[Y=y] \leq \text{Acc}^*_{\mathcal{F}}(X;Y)
\]
for any other random variate $X$ taking values in $\mathcal{X}$.

(iii) If $Y=f(X)$ where $f \in \mathcal{F}$ is a deterministic function, then $\text{Acc}^*_{\mathcal{F}}(X;Y) = 1$.

(iv) For any random vectors $X^{(1)}$ and $X^{(2)}$, the optimized accuracy is greater for the concatenated vector $X = (X^{(1)}, X^{(2)})$ than for $X^{(1)}$ alone,
\[
\text{Acc}^*_{\mathcal{G}}(X;Y) \geq \text{Acc}^*_{\mathcal{F}}(X^{(1)};Y).
\]
\end{theorem}

\noindent\textbf{Proof of theorem.}
\label{proof.acc.theorem}

(i) follows from the fact that accuracy is a probability. (iii) is also trivial to show from the definitions.

Now we will show (ii).
Let $\pi_y =\Pr[Y=y]$ for $y\in \{1,\hdots, k\}$.
We claim that the decoder defined by
\[
f^*(x) = \text{argmax}_{y \in \mathcal{Y}} \pi_y.
\]
is optimal.
To see this, note that
\[
\Pr[f^*(X) = Y] = \Pr[Y = \text{argmax}_{y \in \mathcal{Y}} \pi_y] = \max_y \pi_y.
\]
Meanwhile, for any decoder $f$, note that because $X' \perp Y$, then also $f(X') \perp Y$.  Therefore
\begin{align*}
\Pr[f(X') = Y] &= \sum_{y=1}^k \Pr[(f(X')=y) \cap (Y=y)]
\\&= \sum_{y=1}^k \Pr[f(X')=y] \Pr[Y=y] \text{ (independence) }
\\&= \sum_{y=1}^k \pi_y \Pr[f(X')=y] 
\\&\leq \max_{y=1}^k \pi_y
\end{align*}
where the last line follows from the fact that $\Pr[f(X') = Y]$ is a convex combination over $\pi_y$.

To show (iv), note that given any $\mathcal{F}$-optimal decoder for $X^{(1)}$, $f(x^{(1)})$, one can at least ensure the same accuracy in the class $\mathcal{G}$ by taking a decoder $g$ that is defined as $g(x^{(1)},x^{(2)}) = f(x^{(1)})$. $\Box.$

We now comment on some connections between the accuracy \eqref{eq:Acc}, cross-accuracy \eqref{eq:XAcc} and decoding divergence to information-theoretic quantities \cite{cover2012elements}.

Given a discrete random variate $X$ taking on values in $\mathcal{X}$ with probabilities $(p_x)_{x\in\mathcal{X}}$, the entropy $H(X) = -\sum_x p(x) \log(p(x))$ is obtained as a risk in a forecasting problem.  Suppose we want to predict the distribution of $X$ via an estimated distribution $(q_x)_{x\in\mathcal{X}}$.  Upon observing $X=x$, our cost is $-\log(q_x)$.  Hence, we pay a very large cost if $x$ had a small probability our the estimated distribution.  The best possible prediction we can make (that minimizes \emph{risk} or average cost) is to take $q_x = p_x$ for $x \in \mathcal{X}$.

The risk of predicting $X \sim p$ via an estimated distribution $q$ is the cross-entropy $H(p, q)=-\sum_{x\in\mathcal{X}} p_x \log(q_x)$, and the minimizer of cross-entropy is the true distribution $p$, which achieves $H(p,p) = H(p)$.  Analogously, we consider cross-accuracies between two encodings given by joint distributions $(X, Y)$ and $(X', Y')$.
\[\text{XAcc}^*_{\mathcal{F}}((X;Y) \to (X';Y')) = \Pr[f^*(X') = Y'],\]
where $f^*$ is the optimal decoder for $(X;Y)$.  If we hold the second argument $(X';Y')$ fixed, the encoding that maximizes of the cross-accuracy (over the first argument) is the same encoding $(X';Y')$, which yields
\[
\text{XAcc}^*_{\mathcal{F}}((X';Y') \to (X';Y')) = \text{Acc}^*_{\mathcal{F}}(X';Y').
\]

Hence, the analogical relationship between our $(\text{Acc}^*_{\mathcal{F}}, \text{XAcc}^*_{\mathcal{F}})$ and Shannon's $(H(p), H(p, q))$ is that (i) both our measures and Shannon's measures are derived from risks in prediction problems (although accuracy is the complement of risk under 0-1 loss); (ii) the first measure (accuracy or entropy) is a one-argument function that obtained as a optimum over one of the arguments in the second measure (cross-accuracy or cross-entropy), which is a two-argument function.

In further analogy, we consider the KL divergence is the difference of cross-entropy and entropy.  Similarly, we define a decoding-based measure of divergence as the difference of optimized accuracy and accuracy:

\[\text{D}^*_{\mathcal{F}}((X';Y')||(X;Y)) = \text{Acc}^*_{\mathcal{F}}(X';Y') - \text{XAcc}^*_{\mathcal{F}}((X;Y) \to (X';Y')).\]
The decoding divergence is not symmetric, but rather exists in two different directions (by switching which encoding is used for training), just as KL divergence. Hence, we average divergences in both directions to get the symmetric decoding divergence
\begin{align*}
\text{SD}^*_{\mathcal{F}}((X;Y),(X';Y')) =& \frac{1}{2}(\text{Acc}^*_{\mathcal{F}}(X;Y) + \text{Acc}^*_{\mathcal{F}}(X';Y') \\&- \text{XAcc}^*_{\mathcal{F}}((X;Y) \to (X';Y')) \\&- \text{XAcc}^*_{\mathcal{F}}((X';Y') \to (X;Y))).
\end{align*}
which gives us a better ability to detect differences in encoding compared to using optimized accuracies alone.  This is akin to how KL divergence is better at measuring differences between distributions than the difference of their respective entropies.

We will prove that the decoding divergence and symmetric decoding divergence are non-negative.

\begin{theorem}
\[\text{D}^*_{\mathcal{F}}((X';Y')||(X;Y)) \geq 0\] and
\[\text{SD}^*_{\mathcal{F}}((X;Y),(X';Y')) \geq 0.\]
\end{theorem}

\noindent\textbf{Proof.}
\label{proof.SD}

Let $f^*=\text{argmax}_{f \in \mathcal{F}} \Pr[f(X) = Y]$
and $g^*=\text{argmax}_{f \in \mathcal{F}} \Pr[f(X') = Y']$.
Hence 
\[
\Pr[g^*(X') = Y'] = \text{max}_{f \in \mathcal{F}} \Pr[f(X') = Y'] \geq \Pr[f^*(X') = Y'].
\]
Recall that
\[
\text{Acc}^*_{\mathcal{F}}(X';Y') = \Pr[g^*(X')=Y']
\]
and
\[
\text{XAcc}^*_{\mathcal{F}}((X;Y) \to (X';Y')) = \Pr[f^*(X') = Y'].
\]
Therefore,
\[
\text{D}^*_{\mathcal{F}}((X';Y')||(X;Y)) = \Pr[g^*(X')=Y'] - \Pr[f^*(X') = Y'] \geq 0.
\]
Meanwhile since
\begin{align*}
\text{SD}^*_{\mathcal{F}}((X;Y),(X';Y')) =& \frac{1}{2}(\text{D}^*_{\mathcal{F}}((X';Y')||(X;Y)) \\&+ \text{D}^*_{\mathcal{F}}((X;Y)||(X';Y')))
\end{align*}
is an average of two non-negative terms, we also have $\text{SD}^*_{\mathcal{F}}((X;Y),(X';Y')) \geq 0.$ $\Box$

%\subsection{Consistency}\label{sec:consistency}

\subsection{Theoretical basis for Z-test and combining Z-tests}\label{sec:Z.test.theory}

This section contains theoretical results covering both \S\ref{sec:testing} and the paragraph `Label imbalances in a naturalistic setting' in \S\ref{sec:control.matching}, and therefore should be read after reading both sections.
Here we prove an asymptotic normality of the test statistic $\text{SD}^n$ under the null distribution in the case that $Y(t)$ is not controlled, and hence where covariate matching (described in \S\ref{sec:control.matching} and \S\ref{sec:cm_details}) is applied.  Our theory in this section requires $X(t)$ are to be i.i.d. conditional on $Y(t)$, but we relax this assumption in \S\ref{sec:VIF_details}.

\subsubsection{Modeling assumptions}

Suppose we have collected data $(X^{ij}(t),Y^{ij}(t))$ for $i \in \{1, 2\}$ indexing the two decoders to be compared (e.g. 1=free-running, 2=task), $j\in \{\alpha,\beta\}$ indexing the partition (training or test, see Algorithm \ref{alg:train_test_partition}), and $1 \leq t \leq n_{ij}$ indicating the time index (which could be concatenated over several recordings.)  Note that the data partitioning algorithm (Algorithm \ref{alg:train_test_partition}) generalizes to the case of $n_d$ decoders to be compared, which would allow a test of the effect of a confounding factor on encoding for a confounding factor with more than two levels.  However, here we only analyze the case $n_d=2$.

We will make the following assumptions.  In one of the assumptions, we require the concept of \emph{Markov blanket}.  In any system of random variables, the Markov blanket defines the set of variables which directly influence a given random variate $X$.

\begin{definition}\label{def:MB}
Let $X_1,\hdots, X_m$ be random variables.  Then we say that $\{X_i\}_{i \in I}$ is a \emph{Markov blanket} for $X_j$ with respect to $\{X_\ell\}_{\ell=1}^m$ if and only if
\[
X_i \perp (X_\ell)_{\ell \notin \{j\}\cup I} | (X_i)_{i \in I}.
\]
\end{definition}

\begin{assumption}\label{as:cond_dist}
There exist conditional distributions $f_{i,y}(x)$ for $i \in \{1,2\}$ and $y \in \{1,\hdots, n_c\}$ such that
\[
X^{ij}(t) \sim f_{i, Y^{ij}(t)}
\]
for all $i \in \{1,2\}$, $j\in \{\alpha,\beta\}$, $1 \leq t \leq n_{ij}$.
\end{assumption}

\begin{assumption}\label{as:cond_ind}
Defining $Z^{ij} = (X^{ij}(t), Y^{ij}(t))$, then $\{Z^{ij}\}_{1\leq i \leq 2, j \in \{\alpha, \beta\}}$ are mutually independent.  Furthermore, for any given $(i,j)$ and $1 \leq t \leq n_{ij},$ the Markov blanket of $X^{ij}(t)$ is $\{Y^{ij}(t)\}$ (with respect to $\{Z^{ij}\}_{1\leq i \leq 2, j \in \{\alpha, \beta\}}$).
\end{assumption}

The null hypothesis is where the conditional distributions are the same across the confounding factor, i.e. 
\begin{equation}\label{eq:h0_conditional_dist}
    \text{H0: }f_{1,y} = f_{2,y}\text{ for all }y \in \{1,\hdots, n_c\}.
\end{equation}

Under these assumptions, we want to show that the symmetric decoding divergence is asymptotically normal under the appropriate asymptotic limit.
The limit we need is one where the minimal class count $n_{min}$ goes to infinity, defined as
\begin{equation}\label{eq:n_min}
n_{min} = \text{min}_{1 \leq i \leq 2, j \in\{\alpha,\beta\}, 1\leq y\leq n_c} 
\sum_{t=1}^{n_{ij}} I(Y(t) = y).
\end{equation}

Our test statistic for detecting differences between encodings $(X,Y)$ and $(X',Y')$ is the empirical symmetric decoding divergence, motivated in section \ref{sec:info_theory}.
\begin{align*}
\text{SD}^n((X;Y), (X';Y'))=& \frac{1}{2}(\text{Acc}^n(X; Y) + \text{Acc}^n(X'; Y')
\\&- \text{XAcc}^n((X;Y) \to (X';Y'))\\&- \text{XAcc}^n((X';Y') \to (X;Y))
).
\end{align*}

We will need one more additional technical assumption to show that the symmetric decoding divergence will be asymptotically normal, which is that the optimal decoders for either context are unique\footnote{Otherwise, the symmetric decoding divergence will be a mixture of normal distributions, for which a Z-test might still be appropriate (using an upper bound on the standard deviation over mixture components), but we have not yet formalized this in a rigorous fashion.}, and that the trained decoders converge to the optimal decoder as $n \to \infty$.  

\begin{assumption}\label{as:acc_bound}
Assume that over a sequence of problems where $n \to \infty$, where $n=n_{min}$ as defined in \eqref{eq:n_min}, the assumptions \ref{as:cond_dist} and \ref{as:cond_ind}.
Suppose $(X_{tr}^i(t), Y_{te}^i(t))_{t=1}^{n_{tr}}$ for $i=1,2$ are the training sets and $(X_{te}^i(t), Y_{te}^i(t))_{t=1}^{n_{te}}$ are the test sets obtained by Algorithm \ref{alg:train_test_split}.  
Define for $f \in \mathcal{F}$ the empirical accuracy
\[
\emph{\text{Acc}}^n(f; i) = \frac{1}{n_{te}} \sum_{t=1}^{n_{te}} I(f^i(X_{te}^i(t) = Y_{te}^i(t))
\]
and the true accuracy
\[
\emph{\text{Acc}}^\infty(f; i) = \frac{1}{n_c}\sum_{y=1}^{n_c} \sum_{x \in \mathcal{X}} f_{i, y}(x) I(f(x) = y)\}.
\]
Then, for each $i=1,2$, there exists a unique $f^{i*}$ that optimizes the true accuracy $\emph{\text{Acc}}^\infty(f; i)$.  Also, the trained decoder $f^i$ satisfies
\[
\lim_{n\to\infty} \Pr[f^*=f^{i*}] = 1
\]
for $i=1,2$.
\end{assumption}

In particular, assumption \ref{as:acc_bound} holds if the following are true.
\begin{assumption}\label{as:extra_as}
(i) the decoder $f^i$ is trained by the optimizing the training set accuracy over a function class $\mathcal{F}$; 

(ii) the space of neural activity $\mathcal{X}$ is a finite set, and 

(iii) that every $x \in \mathcal{X}$ has a non-zero probability of appearing in either context, in the sense that
\begin{equation}\label{eq:min_X_prob}
\text{min}_{i=1,2} \text{max}_{y=1}^{n_c} f_{i, y}(x) \geq 0,
\end{equation}
\end{assumption}
as we will show.

\begin{theorem}
Assume that over a sequence of problems where $n \to \infty$, where $n=n_{min}$ as defined in \eqref{eq:n_min}, the assumptions \ref{as:cond_dist}, \ref{as:cond_ind}, and \ref{as:extra_as} hold.
Define the set of optimal decoders $\mathcal{F}^{i*}$ by
\[
\mathcal{F}^{i*} = \{f: \text{Acc}^\infty(f;i) = \emph{\text{max}}_{f \in \mathcal{F}} \text{Acc}^\infty(f;i),
\]
Suppose $(X_{tr}^i(t), Y_{te}^i(t))_{t=1}^{n_{tr}}$ for $i=1,2$ are the training sets and $(X_{te}^i(t), Y_{te}^i(t))_{t=1}^{n_{te}}$ are the test sets obtains by Algorithm \ref{alg:train_test_split}.
Define $f^i$ by
\[
f^i = \emph{\text{argmax}}_{f \in \mathcal{F}} \emph{\text{Acc}}^n(f; i).
\]
Then we have
\[
\lim_{n \to \infty} \Pr[f^i \in \mathcal{F}^{i*}]=1.
\]
Furthermore, if we additionally suppose that $|\mathcal{F}^{i*}| =1$ for $i=1,2$; that is, there are unique optimal decoders $f^{i*}$, then
\[
\lim_{n \to \infty} \Pr[f^i = f^{i*}]=1.
\]
\end{theorem}

\noindent\textbf{Proof}\label{proof.lim} Since we assumed that both $\mathcal{X}$ and $\mathcal{Y}$ are finite sets, there are at most $|\mathcal{Y}|^{|\mathcal{X}|}$ distinct functions $f$ that map from $\mathcal{X}$ to $\mathcal{Y}$.  Let 
\[
\delta^i = \text{min}_{g \notin \mathcal{F}^{i*}} \text{Acc}^\infty(f^{i*}; i) - \text{Acc}^\infty(g; i).
\]
By the Weak Law of Large Numbers \cite{feller1971introduction} and the fact that $\mathcal{F}$ is finite,
\[
\lim_{n \to \infty} \Pr\left[\underbrace{\text{Acc}^\infty(f; i)}_{\text{true accuracy of }f} - \underbrace{\text{Acc}^n(f; i)}_{\text{empirical accuracy of }f} > \frac{\delta^i}{2}\right] = 0.
\]
But in the event that the empirical accuracy and true accuracy differ but no more than $\delta^i/2$, it follows that the $f$ that optimizes the empirical accuracy is an element of $\mathcal{F}^*$.  $\Box$

\subsubsection{Asymptotic normality of symmetric decoding divergence}

\begin{theorem}\label{the:cm_theorem}
Assume that over a sequence of problems where $n \to \infty$, where $n=n_{min}$ as defined in \eqref{eq:n_min}, the assumptions \ref{as:cond_dist}, \ref{as:cond_ind}, and \ref{as:acc_bound} hold.  
Define
\[
\emph{\text{SD}}^n = \frac{1}{2}(\emph{\text{Acc}}(f^1; 1) + \emph{\text{Acc}}(f^2; 2) - 
\emph{\text{Acc}}(f^1; 2) - \emph{\text{Acc}}(f^2; 1)).
\]

Then, under the null hypothesis \eqref{eq:h0_conditional_dist}, there exists some $\sigma^2 > 0$ such that
\[
\sqrt{n}\sigma^{-1}\emph{\text{SD}}^n \xrightarrow{D} Z
\]
where $Z \sim N(0, 1)$.
\end{theorem}

\noindent\textbf{Proof of theorem \ref{the:cm_theorem}}\label{proof.cm_theorem}

Define
\begin{align*}
\text{SD}^{n*} =& \frac{1}{2}(\text{Acc}(f^{1*}; 1) + \text{Acc}(f^{2*}; 2) \\&- 
\text{Acc}(f^{1*}; 2) - \text{Acc}(f^{2*}; 1)).
\end{align*}
We can write
\[
n_{te}\text{SD}^{n*} = \sum_{t=1}^{n_{te}} W_t
\]
where
\begin{align*}
W_t =& \frac{1}{2}\bigg(
I(f^{1*}(X_{te}^1(t)) = Y_{te}^1) +
I(f^{2*}(X_{te}^2(t)) = Y_{te}^2) \\&-
I(f^{1*}(X_{te}^2(t)) = Y_{te}^2) -
I(f^{2*}(X_{te}^1(t)) = Y_{te}^1)
\bigg).
\end{align*}
Hence, by the central limit theorem, there exists $\sigma^2$ such that
\[
\frac{\sqrt{n_{te}}}{\sigma}\text{SD}^{n*} \xrightarrow{D} Z
\]
for $Z \sim N(0, 1).$
But by assumption \ref{as:acc_bound}, we have
\[
\lim_{n\to\infty} \Pr[\text{SD}^n \neq \text{SD}^{n*}] = 0.
\]
Hence $\text{SD}^n$ has the same limiting distribution. $\Box$

\subsubsection{Standard deviation bound and Z-test combination}

As we mentioned in section \ref{sec:testing}, we can upper bound the standard deviation of the symmetric decoding divergence by a weighted sum of the standard deviations of the component accuracies, which we state in the theorem below.

\begin{theorem}\label{the:sd4}
We have
\begin{align*}
\sigma(\text{SD}^n((X;Y), (X';Y'))) \leq& \frac{1}{2}(
\sigma(\text{Acc}^n(X; Y)) + \sigma(\text{Acc}^n(X'; Y')) 
\\&+ \sigma(\text{XAcc}^n((X;Y) \to (X';Y'))) 
\\&+ \sigma(\text{XAcc}^n((X';Y') \to (X;Y)))),
\end{align*}
where $\sigma(\cdot) = \sqrt{\text{Var}(\cdot)}$.
\end{theorem}

This theorem follows trivially from the following lemma.

\begin{lemma}\label{lem:sd}
(Upper bound on standard deviation of a sum of random variables.)
For random variables $Z_1,\hdots, Z_n$,
\begin{equation}\label{eq:combineZ}
\sigma(Z_1+ \cdots + Z_n) \leq \sum_{i=1}^n \sigma(Z_i) .
\end{equation}
\end{lemma}

\noindent\textbf{Proof of lemma}.\label{proof.std.bound.lemma}

The standard deviation $\sigma(\cdot)$ is a generalized $p$-norm \cite{falie_statistical_2010} which satisfies a generalized form of Minkowski's inequality, \cite{nantomah_generalized_2017},
\[
\left\|\sum_{i=1}^n Z_i\right\|_p \leq \sum_{i=1}^n ||Z_i||_p.
\]
$\Box$.

\paragraph{Other reasons to combine tests.}
Another application of lemma \ref{lem:sd} is in combining multiple Z-tests into one, by averaging estimates and averaging estimated standard deviations, as we mentioned in section \ref{sec:control.matching}, under the paragraph `Confounds'.  The following theorem makes this concrete.

\begin{theorem}
Suppose 
\[Z_i \sim N(\mu_i, \sigma^2_i)\]
and
\[\hat{\sigma}_i \geq \sigma_i\]
for $i=1,\hdots, m$.

Define the combined null hypothesis
\[
H_0: \bar{\mu} = 0
\]
where $\bar{\mu} = \frac{1}{m}\sum_{i=1}^m \mu_i$.

Then defining 
\[
\bar{Z} = \frac{1}{m}\sum_{i=1}^m Z_i,
\ \ 
\bar{\sigma} = \frac{1}{m}\sum_{i=1}^m \hat{\sigma}_i,
\]
\[
p_1 = 1-\Phi\left(\frac{\bar{Z}}{\bar{\sigma}}\right),\ \ 
p_2 = 2\left(1-\Phi\left(\frac{|\bar{Z}|}{\bar{\sigma}}\right)\right)
\]
where $\Phi$ is the cumulative distribution function of a standard normal random variate,
then $p_1$ is a $p$-value for the one-sided test against the alternative $H_1: \bar{\mu} > 0$, and $p_2$ is a $p$-value for the two-sided test against the alternative $H_1: \bar{\mu} \neq 0$.

In other words, under $H_0$, we have $\Pr[p_1 \leq \alpha] \leq \alpha$ and $\Pr[p_2 \leq \alpha] \leq \alpha$ for all $\alpha \in [0,1]$.
\end{theorem}

\noindent\textbf{Proof.} \label{proof.other.reasons}
Let $\sigma_{true}=\sigma(\bar{Z})$.  By Lemma \ref{lem:sd}, we have $\bar{\sigma} \geq \sigma_{true}$.  It follows that $\frac{\bar{Z}}{\bar{\sigma}} \leq \frac{\bar{Z}}{\sigma_{true}}$, and also that $\frac{|\bar{Z}|}{\bar{\sigma}} \leq \frac{|\bar{Z}|}{\sigma_{true}}$.
Hence, under $H_0$ where $\E[\bar{Z}] = 0$,
\begin{align*}
\Pr[p_1 \leq \alpha] =& \Pr\left[1-\Phi\left(\frac{\bar{Z}}{\bar{\sigma}}\right) \leq \alpha\right] \\
\leq& \Pr\left[1-\Phi\left(\frac{\bar{Z}}{\sigma_{true}}\right) \leq \alpha\right] \\=& \alpha.
\end{align*}
Similarly,
\begin{align*}
\Pr[p_2 \leq \alpha] =& \Pr\left[1-\Phi\left(\frac{|\bar{Z}|}{\bar{\sigma}}\right) \leq \alpha/2\right]\\
\leq& \Pr\left[1-\Phi\left(\frac{|\bar{Z}|}{\sigma_{true}}\right) \leq \alpha/2\right] \\=& \alpha.
\end{align*}
Thus we have proved the theorem.  $\Box$

This combining can be applied to combine at multiple levels of analysis.  If one has multiple days of recordings, possibly with different sets of neurons, it can be justified to average tests within an animal.  If the effect is extremely weak, one may even average across animals to estimate a group-level effect.  Another application of averaging is within a single hypothesis test.  The result of the Z-test can be random depending on the random seeds used for label match and used for classifier training.  By averaging Z-tests over random seeds, one can reduce the randomness due to random number generation (also called simulation error) as long as one is willing to pay the computational price of repeatedly sampling results, which can be easily parallelized.

\paragraph{Concrete example}

In our experiments in \S\ref{sec:experiments}, the aggregated decoding accuracy is defined as $\obar{\text{acc}}_i = \frac{1}{m \times n} \sum_{k \in D} \sum_{j=1}^{n} \text{acc}_{i}^{k_j}$, while the aggregated standard error is defined as $\bar{\sigma}_i = \frac{1}{m \times n} \sum_{k \in D} \sum_{j=1}^{n} \sigma_{i}^{k_j}$, where $k$ is the factor level of the confound (forward vs. backward direction) out of the set $D$ of possible directions, $i$ is the context of the test set (same- or cross-context), $k_j$ is the decoder from the $j$th seed, $m$ is the number of decoders to aggregate, and $n$ is the number of random seeds to average over. The one-sided $p$-value is calculated with $p = 1 - \Phi \left( \frac{ \obar{\text{acc}}_{\text{same}} - \obar{\text{acc}}_{\text{cross}} }{ \bar{\sigma}_{\text{same}} + \bar{\sigma}_{\text{cross}}}   \right)$ where $\Phi$ is the c.d.f. of the standard normal distribution.

\subsection{Train Test Partitions}\label{sec:partition}

To preserve the validity of our testing procedure in the face of possible dependence, we first ensure that the training set and the test set are completely independent from each other. We do this by splitting the data from an animal into multiple sub-datasets that can be considered as independent of one another, e.g. multiple recording sessions, or segments of a single long recording. We then create training and test sets by concatenating these while excluding segments that would be between train and test sets; the algorithm is detailed below and yields the desired ratio between training and test examples. We then apply covariate matching, as described earlier, to obtain the final training and test sets for each context. The dataset properties are shown in Table \ref{tab:dataset_specs} below.

\begin{table*}[htb]
    \centering
    \caption{Dataset properties for the 2 method development and 10 held-out mouse sessions. Task data were naturally divided into subdatasets, per trial, and FR data were split into two subdataset halves. "Prop. correct" denotes the proportion of trials in which the subject completed the task correctly.}
    \begin{tabular}{@{\extracolsep{4pt}} lccccccc@{}}
    \\
    \toprule
    
    &\multicolumn{4}{c}{Task Subdatasets} & \multicolumn{3}{c}{FR Subdatasets} \\ 
    \cline{2-5} \cline{6-8}
    \thead{mouse/sess.\\ (\# neurons)} & Counts & \thead{Prop. \\ correct} & \thead{Med. length \\ (min:sec)}  & \thead{Tot. time \\ (min:sec)} & Counts & \thead{Med. length \\ (min:sec)} & \thead{Tot. time \\ (min:sec)} \\
    \cmidrule{1-8}
    37/1 (41)	& 23 & 0.87 & 0:47	& 19:13 & 2 & 9:02	& 18:05 \\
    37/2 (38)	& 24 & 0.96 & 0:42	& 21:43 & 2 & 12:00	& 24:00 \\
    \\
    36/1 (72)	& 23 & 0.87 & 0:42	& 20:47 & 2 & 10:35	& 21:10 \\
    36/2 (96)	& 24 & 0.96 & 0:50	& 28:46 & 2 & 11:41	& 23:22 \\
    38/1 (52)	& 22 & 0.91 & 0:45	& 17:13 & 2 & 9:10	& 18:20 \\
    38/2 (51)	& 21 & 0.95 & 1:04	& 29:01 & 2 & 11:10	& 22:20 \\
    39/1 (42)	& 21 & 1.00 & 0:42	& 16:10 & 2 & 9:34	& 19:09 \\
    40/1 (104)	& 25 & 0.96 & 0:43	& 24:33 & 2 & 12:14	& 24:28 \\
    43/1 (45)	& 21 & 1.00 & 1:28	& 38:20 & 2 & 11:20	& 22:40 \\ %bad FR dataset.
    44/1 (42)	& 22 & 0.86 & 0:57	& 24:19 & 2 & 9:34	& 19:08 \\
    46/1 (56)	& 19 & 0.89 & 0:49	& 19:00 & 2 & 9:27	& 18:55 \\
    46/2 (57)	& 26 & 1.00 & 0:50	& 25:25 & 2 & 12:11	& 24:22 \\
    \bottomrule
    \end{tabular}

    \label{tab:dataset_specs}
\end{table*}

For each decoder $D$, we subsample its dataset $Z_S^D$, which contains corresponding inputs $X^D$ and targets $Y^D$, to form two partitions $Z_\alpha^D$ and $Z_\beta^D$, in order to maintain independence between train and test data. This is done by randomly selecting subdatasets $S$ to most closely match the desired proportion of data in each partition, $p_\alpha$ and $p_\beta$. For our applications, we set $p_\alpha = p_\beta = 0.5$ to ensure that there is enough data for both train and test. Algorithm \ref{alg:train_test_partition} is summarized below.

\begin{enumerate}
    \item \textbf{Generating Candidate Partitions}: 
    First, we randomly permute subdatasets $S$ to get $\tilde{S}$. Each element of $\tilde{S}$ represents a candidate partition with two complementary subsets of the data. Subset $\tilde{S}_a$ will contain the first $i$ subdatasets, while subset $\tilde{S}_b$ will contain the remaining subdatasets. Random permutation of $S$ is necessary so that we can obtain different partitions given different random seeds.
    
    \item \textbf{Scoring Candidate Partitions}: Candidate partitions are scored by computing the difference between calculated partition proportions and desired partition proportions, $p_a - p_\alpha$. This is done by first storing class counts for each partition are stored into $\vec{n}_a$ and $\vec{n}_b$ which have dimensions $1\times n_c$ where $n_c$ is the number of classes and calculating the partition proportion by dividing the size of one partition by the total size of both partitions. In this context, size is defined as the minimum label counts across all label classes since the effective amount of data is determined by the size of the smallest class rather than the raw number of datapoints. Finally, partition scores are stored into list $e$ for the next step. 
    
    \item \textbf{Partition selection}:
    We find the partition which has the smallest positive score to divide our data into two parts with proportions approximately equal to the desired ones. We find the index $x$ of the subset partition with the smallest positive score. By looking at only the positive scores, we ensure that $p_a \geq p_\alpha$ which is useful in the case where $p_\alpha=p_\beta=0.5$ and we want to reserve slightly more data for the train partition than the test partition if possible. The data is divided into two partitions by assigning subdatasets $\tilde{s}[1,\dots,x]$ to $Z_\alpha$ and the complementary subdatasets $\tilde{s}[x+1,\dots,n_s]$ to $Z_\beta$. 

\end{enumerate}

% Alg 1
\begin{algorithm*}[!htb]
\caption{Partition data Z into two parts}\label{alg:train_test_partition}
\hspace*{\algorithmicindent} \textbf{Input}: Dataset $Z_S^D$ for decoder $D$, which contains corresponding inputs $X^D$ and target labels $Y^D$, with subdatasets $S$, number of target classes $n_c$, and desired proportion of data to be reserved for one partition $p_{\alpha}$.   \\
\hspace*{\algorithmicindent} \textbf{Output}: Two partitions of the original data, $Z_\alpha^D$ and $Z_\beta^D$, that most closely achieves the desired fraction of training data $p_{\alpha}$.
\begin{algorithmic}[1]
\Procedure{splitData}{}
\State $n_S \gets \text{length}(S)$
\State $\tilde{S} \gets $ permute$(S)$ \Comment{see \S\ref{sec:subroutines}}
\State $e \gets [e_x], x = 1, \dots, n_S$ \Comment{Initiate list to store proportion scores}
\For{$i = 1, \dots, n_S$} \Comment{Loop through possible train test splits}
    \State $\tilde{S}_a \gets \tilde{S} [1, \dots, i]$
    \State $\tilde{S}_b \gets \tilde{S} \setminus{\tilde{S}_a}$
    %% get target class counts
    \State $\vec{n}_a \gets [n_{a, x}], x= 1,\dots, n_c$ \Comment{Init list to store label counts}
    \State $\vec{n}_b \gets [n_{b, x}], x= 1,\dots, n_c$
    \For{$j = 1, \dots, n_{c}$} 
        \For{$k \in \{a,b\}$}
            \State $\vec{n}_\text{k, j}  \gets \sum_{S,t} \textbf{I}\left( Y_{S_k}^D(t)=j \right)$ \Comment{get target class counts}
        \EndFor
    \EndFor
    \State $p_a \gets \text{min}(\vec{n}_a) / ( \text{min}(\vec{n}_b)+\text{min}(\vec{n}_a))$ \Comment{Proportion of $a$ split}
    \State $e_i \gets p_a - p_{\alpha}$
\EndFor
\State $x \gets \argmin_{i \in \{1, \dots, n_S \}}(e_i)$ subject to $e_i \geq 0$ \Comment{get split with $p_a$ closest to $p_{\alpha}$}
\State $Z_\alpha^D \gets Z_{\tilde{S}[1, \dots, x]}^D$
\State $Z_\beta^D \gets  Z_{\tilde{S}[x+1, \dots, n_s]}^D$
\State \textbf{return} $Z_\alpha^D, Z_\beta^D$ 
\EndProcedure
\end{algorithmic}
\end{algorithm*}

\subsection{Covariate Matching}\label{sec:cm_details}

\paragraph{Description}

We perform covariate matching on the set of all train and test partitions from all decoders, $Z_\alpha$ and $Z_\beta$, to account for confounding dataset properties such as differences in dataset sizes and label distributions. 
We borrow a method from causal inference called covariate matching, which is appropriate because the problem of comparing accuracies between contexts can be expressed in the language of potential outcomes: what is the average causal effect of context $F$ on the the error of a given test example $E$? 
Partitions are subsampled using algorithm \ref{alg:train_test_split} to get the train and test splits, $Z_\text{tr}$ and $Z_\text{te}$, which are used to build the logistic regression models. Covariate matching procedures are different between train and test splits due to their different goals and constraints. For the train data, we maximize model performance by first matching label distributions across the four train splits by randomly subsampling timepoints without replacement, and then oversampling the matched distribution to optimize for balanced accuracy during prediction. For the test data, randomly subsample timepoints in a way that include only unique timepoints so that we can later apply time series methods to estimate the standard error. Algorithm \ref{alg:train_test_split} is summarized below.

\begin{enumerate}
    \item \textbf{Obtaining Class Counts}: We first obtain label class counts for all train and test partitions, and store them into matrices $c_\alpha$ and $c_\beta$ which have dimensions $n \times n_c$. Each element $c_{i,xy}$ represents the sample count for target class $y$ from dataset $x$ of the $i$th partition. 
    
    \item \textbf{Covariate Matching for Train Splits}: For each train partition in $Z_\alpha$, we match label distributions creating a list of timepoints $T_\alpha^{D_i}$ for the $i$th dataset. We randomly subsample without replacement by drawing $m_j$ number of samples where $m_j$ is defined as the minimum class count across all datasets for the $j$th class. We then optimize for balanced accuracy by applying a random oversampler to the matched data. Oversampling is done by randomly drawing samples from the minority classes with replacement until their target counts match the counts of the majority class. 
    
    \item \textbf{Covariate Matching for Test Splits}:
    For each train partition, we match label distribution and optimize for balance accuracy in a single step. We generate test split timepoints $T_\beta^{D_i}$ from partition $Z_\beta$ by randomly drawing without replacement $m$ samples for all target classes, where $m$ is the minimum of $c_\beta$. By drawing the same number of samples for each class across all of the datasets, we end up with test splits which have $3m$ unique samples and a uniform label distribution. Timepoints within each test split must be sorted from least to greatest so that we can estimate variance using the prediction error autocorrelation.
    
    \item  \textbf{Obtaining Train and Test Splits}:
    Using generated timepoints $T_\alpha^{D}$ and $T_\beta^D$, we obtain the set of train and test splits for all datasets, $Z_\text{tr}$ and $Z_\text{te}$. 
\end{enumerate}

\paragraph{Theory}

Please see \S\ref{sec:Z.test.theory} for a conditions and proof of the asymptotic normality of the test statistic obtained by using our method with covariate matching.

% Define
% \[
% W_i = \frac{1}{2}\left(I(f^i())\right)
% \]

%\paragraph{Pseudocode}

% Alg 2
\begin{algorithm*}[!htb]
\caption{Create train test splits with balanced dataset properties}\label{alg:train_test_split}
\hspace*{\algorithmicindent} \textbf{Input}: Two sets of partitions for all $n_d$ datasets from every factor level, $Z_\alpha$ and $Z_\beta$, where $Z_S = \{Z^{D_i}_\text{S} | 1\leq i\leq n_d, i \in \bb{Z}\}$, $n_c$ is the number of classes. \\
\hspace*{\algorithmicindent} \textbf{Output}: Train and test splits for all datasets, $Z_\text{tr}$, $Z_\text{te}$, accounting for dataset property confounds
\begin{algorithmic}[1]
\Procedure{TrainTestSplitFromPartitions}{}
\State $c_\alpha \gets [c_{xy}], x = 1, \dots, n_d, y = 1,\dots, n_c$ \Comment{Init list to store label counts}

% Get train/test counts
\State $c_\beta \gets [c_{xy}], x = 1, \dots, n_d, y = 1,\dots, n_c$
\For{$i \in \{\alpha, \beta \} $}
    \For{$j = 1,\dots, n_d $}
        \State $c_{i, j, k \in \{1, \dots, n_c\}} \gets  \text{getCountsPerTargetClass}(Z_i^{D_j})$ \Comment{see \S\ref{sec:subroutines}}
    \EndFor
\EndFor

% Balance Train split

\State $Z_{\text{tr}} \gets \{ Z_{\alpha}^{D_i}(T_{\alpha}^{D_i}) |  1 \leq i \leq n, i \in \bb{Z} \} $
\For{$i = 1,\dots, n_d$} \Comment{Match train label distributions}
    \State $t_\alpha \gets [t_x], x=1,\dots, \sum_{y=1}^{n_c} c_{\alpha,iy}$
    \For{$j = 1, \dots, n_c$}
        \State $m_j \gets \min_{k \in \{1, \dots, n_d\}} (c_{\alpha, kj})$ \Comment{Take min across all datasets for class $j$}
        \State $t_\alpha[m_{j-1}, \dots, m_{j}] \gets $ draw($Z^{D_i}_{\alpha}$, $m_j$, $j) \text{, where } m_0 = 1$ \Comment{see \S\ref{sec:subroutines}}
    \EndFor
    \State $T_{\alpha}^{D_i} \gets \text{oversample}(t_\alpha)$ \Comment{see \S\ref{sec:subroutines}}
\EndFor

% Balance Test split

\State $Z_{\text{te}} \gets \{ Z_{\beta}^{D_i}(T_{\beta}^{D_i}) |  1 \leq i \leq n, i \in \bb{Z} \} $
\State $m \gets \min (c_{\beta})$
\For{$i = 1,\dots, n_d$}  \Comment{Draw unique test samples}
    \State $t_\beta \gets [t_x], x=1,\dots, 3m$
    \For{$j = 1, \dots, n_c$}
        \State $t_\beta[m(j-1)+1, \dots, mj] \gets $ draw($Z^{D_i}_{\beta}$, $m$, $j$)
    \EndFor
    \State $T_{\beta}^{D_i} \gets \text{sort}(t_\beta)$ \Comment{Timepoints are sorted to calculated ACC later. See \S\ref{sec:subroutines}}
    
\EndFor

\State \textbf{return} $Z_{\text{tr}}$, $Z_{\text{te}}$
\EndProcedure
\end{algorithmic}
\end{algorithm*}

\subsection{Correlated data and VIF estimation}\label{sec:VIF_details}

\subsubsection{Model for neural data with correlations}

Our theory depends on a strong assumption on the stochastic processes $Y(t)$, $\epsilon(t)$ called \emph{limited dependence range}.  We define it as follows.

\begin{definition}
Let $Z(t), t=1,2,\hdots$ be a stationary stochastic process \cite{feller1971introduction} with the following property: there exists some integer $k > 0$ such that for any set of times $1 \leq t_1 \leq \hdots \leq  t_\ell \leq t_{\ell+1} \leq \cdots \leq t_m$ such that $(t_{\ell + 1} - t_\ell) \geq k$, we have
\[
(Z(t_1),\hdots, Z(t_\ell)) \perp (Z(t_{\ell+1}),\hdots, Z(t_m)).
\]
Then we say that $Z(t)$ has \emph{dependence range} $k$.
\end{definition}

We believe that it is plausible that neural data, when suitably pre-processed, approximately satisfies this condition for some suitable range $k$.

Our model for the data is given by the following assumption.  

\begin{assumption}\label{as:limited_dependence}
Let $Y(t)$ be a stationary time series for the target $Y$, for time steps $t=1,\hdots$, such that the label counts are asymptotically equal (this can be achieved by Algorithm \ref{alg:train_test_split})
\begin{align*}
\lim_{n \to \infty} \frac{1}{n} \sum_{t=1}^n I(Y(t)=1) =& \cdots \\=& \lim_{n \to \infty}\frac{1}{n} \sum_{t=1}^n I(Y(t)=n_c) = \frac{1}{n_c}.
\end{align*}
Let $\epsilon(t)$ be a times series of error variates taking values in $\mathbb{R}^q$.  
Let $f_{y, \epsilon}(x)$ be a parametric family of distributions over the space of neural signals $\mathcal{X}$, and assume that $X(t)$ is drawn from the distribution $f_{Y(t),\epsilon(t)}$ independently of all other neural signals $X(u)$ for $u \neq t.$
Further assume that the stochastic process $\epsilon(t)$ is independent of the process $Y(t)$.
(From the stationarity of $Y(t)$ and $\epsilon(t)$ we can also conclude that $X(t)$ is stationary.)
Assume further that there exists some integer $0 < k < \infty$ such that both $\epsilon(t)$ and $Y(t)$ (and, by extension, $X(t$) has dependence range $k$.)
\end{assumption}

If we have an lower bound $\hat{k}$ on this integer $k$, then we can consistently estimate the standard deviation of $\text{Acc}^n$ conditional\footnote{The same result applies for the \emph{unconditional} standard deviation if we assume an analogue of Assumption \ref{as:acc_bound}.} on the trained decoder $f$, as we state in the following result.  For this reason, we call such an estimate $\hat{k}$ an estimated variance inflation factor (VIF).

\begin{theorem}
(Consistent estimation of decoder-specific standard deviation given known VIF.)
Assuming assumption \ref{as:limited_dependence}, let $f$ be a fixed, trained decoder.
Let $E(t) = I(f(X(t)) = Y(t))$, and define $\emph{\text{Acc}}^n = \frac{1}{n}\sum_{t=1}^n E(t).$

Given $\hat{k} \geq k$, define the following estimator of $\sigma(\text{Acc}^n|f)$,
\begin{equation}\label{eq:hat_sigma}
\hat{\sigma}^n = \sqrt{\frac{\hat{k}}{n^2} \sum_{t=1}^n (E(t) - \bar{E})^2 },
\end{equation}
where $\bar{E} = \frac{1}{n}\sum_{t=1}^n E(t).$

Then $\hat{\sigma}^n$ is asymptotically an upper bound, in the following sense.  For any $\epsilon > 0$, we have
\[
\lim_{n\to\infty} \Pr\left[\frac{\hat{\sigma}^n}{\sigma(\text{Acc}^n|f)} > 1+\epsilon\right] = 0.
\]
\end{theorem}

\noindent\textbf{Proof}\label{proof.upper.bound}
Due to stationarity of both $Y(t)$ and $\epsilon(t)$, we can compute 
the expected accuracy as
\[
\mu = \frac{1}{n_c}\sum_{y=1}^{n_c} \int_{\mathbb{R}^q} \int_{\mathcal{X}} I(f(x) = y) f_{y, \epsilon}(x) dx d\eta(\epsilon) 
\]
where $\eta$ is the marginal distribution of $\epsilon(t).$
It follows from the stationarity of $(X(t), Y(t))$ that $E(t)$ is also stationary.  Furthermore, $E(t)$ is Bernoulli with mean $\mu$ and standard deviation
\[
\sigma(E(t)) = \sqrt{\mu(1-\mu)}.
\]
By the weak law of large numbers, we have the following convergence in probability as $n\to \infty$.
\[
\sqrt{\frac{1}{n} \sum_{t=1}^n (E(t) - \bar{E})^2 } \xrightarrow{p} \sqrt{\mu(1-\mu)}.
\]
Therefore, to prove our claim, it suffices to show that the true standard deviation satisfies
\[
\lim_{n\to\infty} \sqrt{n}\sigma(\text{Acc}^n|f) \leq \sqrt{k} \sqrt{\mu(1-\mu)}.
\]

We will first show this for $n=(m+1)k$ for some integer $m$.  We have
\begin{align*}
n\text{Acc}^n =& E_1 + \cdots + E_n
\\=& (E_1 + E_{k+1} + \cdots E_{mk+1}) 
\\&+ (E_2 + E_{k+2} + \cdots E_{mk+2}) + \cdots 
\\&+ (E_k + E_{2k} + \cdots E_{(m+1)k})
\\=& T_1 + \cdots + T_k
\end{align*}
where $T_i = E_i + E_{k+i} + \cdots E_{mk+i}$ for $i = 1,\hdots, k$.  (We have used the notation $E_i = E(i)$ to save space.)
Crucially, due to the dependence range, the summands comprising $T_i$ are independent, hence
\[
\sigma^2(T_i) = \sigma^2(E_i) + \cdots + \sigma^2(E_{mk+i}) = (m+1)\mu(1-\mu).
\]
Hence $\sigma(T_i) = \sqrt{m+1}\sqrt{\mu(1-\mu)}$.
Now, using Lemma \ref{lem:sd}, we have
\begin{align*}
\sigma(n\text{Acc}^n|f) &= \sigma(T_1 + \cdots + T_k) 
\\&\leq \sigma(T_1) + \cdots + \sigma(T_k) 
\\&= k \sqrt{m+1}\sqrt{\mu(1-\mu)} = \sqrt{nk}\sqrt{\mu(1-\mu)}.
\end{align*}
Dividing both sides by $\sqrt{n}$, we have $\sqrt{n}\sigma(\text{Acc}^n|f) \leq \sqrt{k} \sqrt{\mu(1-\mu)}$.

For $n$ which is not a multiple of $k$, defining $m = \text{ceiling}(n/k)$, we have by similar reasoning that
\[
m\mu(1-\mu) \leq \sigma^2(T_i) \leq (m+1)\mu(1-\mu).
\]
and hence
\[
\sqrt{n}\sigma(\text{Acc}^n|f) \leq k \sqrt{\frac{m+1}{n}}\sqrt{\mu(1-\mu)}.
\]
But since
\[
\lim_{n\to\infty} \frac{n}{m} =\lim_{n\to\infty} \frac{n}{m+1} =  k,
\]
we get
\[
\lim_{n\to\infty} \sqrt{n}\sigma(\text{Acc}^n|f) \leq \sqrt{k} \sqrt{\mu(1-\mu)}
\]
in the case of general $n$. $\Box$

\subsubsection{VIF estimation}

\paragraph{Description}

As we described in section \S\ref{sec:control.matching} and illustrated in Figure \ref{fig:autocov}, our VIF procedure is based on finding an empirical lag $\hat{k}$ where the autocorrelation crosses zero.  This heuristic is, however, vulnerable to the presence of negative correlations within the true autocorrelation curve.  Possible causes of negative autocorrelation are high-frequency noise in the raw signal (prior to spike sorting) and also spike bursts.  High frequency noise (e.g. commonly assumed to be $\geq 250 Hz$) is effectively neutralized by taking bins that are much longer than the noise wavelength.  For example, for our data, the bin is 40ms while the wavelength of 250 Hz noise is 4 ms.  Meanwhile, spike bursts are known in the literature to last as long as 36 ms \cite{harris_temporal_2001}.  Within the maximum span of a spike burst, we may see negative correlations.  A conservative approach to account for these possible negative correlations is to set an integer $k_{min}$ that is greater than the maximum temporal extent of a spike burst, as measured in bins.  This $k_{min}$ should be set based on prior knowledge.  Our VIF procedure (Algorithm \ref{alg:vif}) estimates $k$ based on the autocorrelation, but ignores the autocorrelation curve below $k_{min}.$  In our analysis in \S\ref{sec:experiments}, we took $k_{min}=1$ since a single bin is already longer than our prior estimate for the spike burst length.

We analyze our VIF estimation procedure (Algorithm \ref{alg:vif}) in two ways.
To establish conditions under which it yields a conservative (overestimate) of the variance, we use a theoretical analysis.
To show that it obtains reasonable estimates under realistic conditions, we use a simulation study.

\paragraph{Pseudocode}

Algorithm \ref{alg:vif} estimates the variance inflation factor (VIF) by finding the first negative index of the autocorrelation of the prediction error.

\begin{algorithm*}[!htb]
\caption{Estimate VIF from decoder prediction}\label{alg:vif}
\hspace*{\algorithmicindent} \textbf{Input}:  Minimal VIF $k_{min}$, decoder prediction $\hat{Y}$ and ground truth $Y(t)$\\
\hspace*{\algorithmicindent} \textbf{Output}: Variance inflation factor, VIF
\begin{algorithmic}[1]
\Procedure{GetVIF}{}
\State $E \gets \textbf{I}\left(\hat{Y}(t) \neq Y(t)\right)$ \Comment{see \S\ref{sec:subroutines}}
\State $\bar{E} \gets \text{mean}(E)$
\State $T \gets \text{length}(E)$
\State $\gamma = [\gamma_1, \dots, \gamma_T]$ \Comment{Autocovariance Curve}
\For{$i = k_{min},\dots, T$}{}
    \State $\gamma_i \gets \frac{1}{T-i} \sum_{t=1}^{T-i} (E_{t+i}-\bar{E}) (E_{t}-\bar{E})$
\EndFor
\State VIF $\gets \text{min}\{i: i\geq k_{min}, \gamma_i \leq 0\}$
\State \textbf{return} VIF
\EndProcedure
\end{algorithmic}
\end{algorithm*}

\paragraph{Theory}

Under assumption \ref{as:limited_dependence}, there exists a true mean $\mu = \E[E(1)]$ and
true autocovariance curve
\[
C_i = \text{Cov}(E(1), E(1+i)).
\]
Furthermore, as a consequence of limited dependence range, $C_i = 0$ for all $i \geq k.$
In order for the VIF estimation (Algorithm \ref{alg:vif}) to asymptotically recover $\hat{k} \geq k$, we need the following additional assumption.

\begin{assumption}\label{as:no_cov_0}
For $k_{min} \leq i\leq k$, the following is true:
\[
C_i = \emph{\text{Cov}}(E(1), E(1+i)) \geq 0.
\]
\end{assumption}

We will sketch out a proof that under such an assumption,
the expected value of $\hat{k}$ is typically a little bit larger than $k$.
We omit the theorem statement and proof as it requires many additional technical details,
and we have not yet developed a rigorous theory.

We intuit that as the empirical autocorrelation $\gamma_i$ converges to the true autocorrelation $C_i$, we have
\[
\Pr[C_i \leq 0] \to 0
\]
for $k_{min} \leq i \leq k-1$, hence
\[
\Pr[\hat{k} \geq k] \to 1.
\]
Beyond this, to understand the behavior of $\hat{k}$, we will have to apply the multivariate central limit theorem to the vector $(E(t)E(t+k), E(t)E(t+k+1),\hdots, E(t)E(t+\ell))$ for some small integer $\ell$.  Using multivariate CLT, we can conclude that $\sqrt{n}(\gamma_k,\hdots, \gamma_\ell)$ has an asymptotically normal distribution centered around zero.  This allows us to compute the probability the event that $\text{min} \{\gamma_k,\hdots, \gamma_\ell\} \leq 0$ as well as the expected value of $\hat{k}$ under that event.  Taking suitably large $\ell$, the probability of said event goes to 1 and we obtain the limiting expectation of $\hat{k}.$

\subsection{Alternative hypothesis tests}
\label{sec:alternative.hypothesis.tests}
We compare our approach with alternative univariate and multivariate methods commonly used in two-sample testing. To do so, we assume that neuron spiking activity is conditionally independent given the location (i.e., $X_t \mid Y_t = y \sim \text{Poisson}(\lambda_y)$ for neurons $X$, location $Y$, and location-dependent intensity $\lambda_y$). This allows us to apply two-sample hypothesis tests to evaluate the distribution of firing rates \textit{at each location} to obtain a global estimate for differences in encoding. If encoding does not differ between contexts, then task and free-running firing rates will be drawn from the same distribution at each location. So, if we can reject the null hypothesis that firing rates are drawn from the same distribution \textit{at any of the locations}, we can reject the null hypothesis that encoding does not differ between contexts. 

We apply univariate hypothesis tests (see \S\ref{sec:univariate.HTs}) by testing the firing rates of each neuron at each location. If, after correcting for multiple comparisons, there exists a neuron $i$ in location $y$ with a significant difference in firing rate distribution, we reject the null hypothesis that there is no difference in encoding between contexts. For multivariate hypothesis tests (see \S\ref{sec:multivariate.HTs}), rather than test each neuron in each location, we test the multivariate distribution of firings for all neurons in each location. If this distribution is significantly different between task and free-running trials at any location, after correcting for multiple comparisons, we reject the null hypothesis that encoding does not differ between contexts. In our evaluation of the alternative hypothesis tests on the experimental data (but not the simulations), we also stratify firing rates by movement direction (forward and backward) in addition to location, to be consistent with the approach in our primary analysis. We choose the conservative Bonferroni correction \cite{bland1995multiple} to correct for multiple comparisons. If we let $\{ p_1, p_2, ..., p_n \}$ denote the $p$-values from $n$ hypothesis tests, the Bonferroni-corrected $p$-value is $p_\text{corrected} = n \times \min \{ p_1, p_2, ..., p_n \}$.

To formalize our notation in the sections that follow, we let $X = (X_t)_{t=1}^T = (X_t^1, X_t^2, ..., X_t^P)_{t=1}^T$ and $X' = (X'_t)_{t=1}^{T'} = (X_t^{'1}, X_t^{'2}, ..., X_t^{'P})_{t=1}^{T'}$ denote the firings of $P$ neurons in $T$ task and $T'$ free-running trials, respectively. The corresponding locations are denoted $y = (y_t)_{t=1}^T$ and $y' = (y'_t)_{t=1}^{T'}$, where $y, y' \in Y$ (in this case, $Y = \{ 1, 2, 3 \}$); the movement direction is denoted $d = (d_t)_{t=1}^T$ and $d' = (d'_t)_{t=1}^{T'}$, where $d \in D$ (here, $D = \{1, 2 \}$, indicating forward or backward movement). To stratify by location (in the case of simulation data) or by location and movement direction (in the case of experimental data), we let $Z_t$ and $Z'_t$ denote the stratification values for task and free-running trials, respectively, at time $t$. Formally, we let $Z_t = y_t, Z'_t = y'_t$ in the simulations, and $Z_t = (y_t, d_t), Z'_t = (y'_t, d'_t)$ in the experimental data analysis. We denote the set of levels by which to stratify the firing rates with by $C$, where $C = Y$ in the simulations and $C = \{ (y, d) \mid y \in Y, d \in D \}$ in the experimental data.

\subsubsection{Univariate hypothesis tests}
\label{sec:univariate.HTs}
We consider three univariate tests to compare the firing rates of individual neurons: an independent $t$-test to compare mean firings, and Kolmogorov-Smirnov and chi-squared tests to compare the distribution of firings. We let $\text{HT}_{UV}(X_1, X_2)$ denote a univariate hypothesis test that takes two univariate samples as inputs and returns a $p$-value representing the probability of the observed values, under the null hypothesis that there is no difference between $X_1$ and $X_2$. If there is a significant difference in any neuron at any stratification level between task and free-running firing rates after correcting for multiple comparisons, then we conclude encoding is significantly different between contexts. We detail this approach formally in Algorithm \ref{alg:univariate.alternative.HT}.

\begin{algorithm*}[!htb]
\caption{Univariate alternative hypothesis tests}\label{alg:univariate.alternative.HT}
\hspace*{\algorithmicindent} \textbf{Input}: task firing rates $(X_t^1, X_t^2, ..., X_t^P)_{t=1}^T$ and stratification values $(Z_t)_{t=1}^T$, free-running firing rates $(X_t^{'1}, X_t^{'2}, ..., X_t^{'P})_{t=1}^{T'}$ and stratification values $(Z'_t)_{t=1}^{T'}$, stratification levels $C$, and hypothesis test $\text{HT}_{UV}$ \\
\hspace*{\algorithmicindent} \textbf{Output}: $p$-value for the difference in context using alternative hypothesis test $\text{HT}_{UV}$, with Bonferroni correction for multiple comparisons
\begin{algorithmic}[1]
\Procedure{UnivariateAlternativeHypothesisTest}{}
\State $\text{pvalues} = [ \, ]$ \Comment{Init list with uncorrected $p$-values. See \S\ref{sec:subroutines}}
\For{$p = 1, \dots, P$} 
    \For{$c \in C$} \Comment{}
        \State $X^{p, c} = [ \, ]$
        \State $X^{'p, c} =  [ \, ]$
        
        \For{$t = 1, \dots, T$}
            \If{$Z_t = c$}
                \State $X^{p, c} \gets \text{append}(X^{p, c}, X_t^p)$ \Comment{task neuron $i$, stratification level $c$}
            \EndIf
        \EndFor
        
        \For{$t = 1, \dots, T'$}
            \If{$Z'_t = c$}
                \State $X^{'p, c} \gets \text{append}(X^{'p, c}, X_t^{'p})$ \Comment{free-running neuron $i$, stratification level $c$}
            \EndIf
        \EndFor
        
        \State $\text{pvalues} \gets \text{append}(\text{pvalues}, \text{HT}_{UV}(X^{p, c}, X^{'p, c}))$
    \EndFor
\EndFor
\State \textbf{return} $P \times \text{length}(C) \times \min (\text{pvalues})$ \Comment{Bonferroni correction}
\EndProcedure
\end{algorithmic}
\end{algorithm*}

\paragraph{Student's \textbf{\textit{t}}-test}
The first univariate test we consider is the two-sample $t$-test, which we use to compare  the mean firing rates of neuron $i$ between contexts (i.e., whether the mean of $X^i$ is significantly different from $X^{i'}$). Using $\overline{X^i}$ and $\overline{X^{'i}}$ to denote the mean of $X^i$ and $X^{'i}$, respectively, the test statistic for the $t$-test is
\begin{equation*}
    t = \frac{\overline{X^i} - \overline{X^{'i}}}{\sqrt{
    s^2(\frac{1}{T} + \frac{1}{T'})}},
\end{equation*}
where $s^2$ denotes the pooled sample covariance and is defined as
\begin{equation*}
    s^2 = \frac{\sum_{j=1}^T(X_j^i - \overline{X^i})^2 + \sum_{k=1}^{T'}(X_k^{'i} - \overline{X^{'i}})^2}{T + T' - 2}.
\end{equation*}
Under the null hypothesis, the test statistic follows a $t$-distribution with $T + T' - 2$ degrees of freedom. 
In addition to assuming the data are independent and identically distributed (i.i.d.), the $t$-test assumes $X$ and $X'$ are continuous and normally distributed, with the same variance \cite{lowry2014concepts}.

\paragraph{Kolmogorov–Smirnov}
We use a Kolmogorov-Smirnov (KS) test to evaluate whether $X$ and $X'$ are drawn from the same probability distribution. The two-sample KS test statistic is defined as the maximum difference between the empirical distribution functions of two distributions. We let $F_{X^i}(r)$ and $F_{X'^i}(r)$ denote the empirical distribution function of the firing rates, $r$, of neuron $i$ in task and free-running trials, respectively. The empirical distribution function is defined as $F(r) = \frac{1}{n} \sum_{i=1}^n \mathbf{I} (x_i \leq r)$. The KS test statistic is then
\begin{equation*}
    \text{KS} = \sup_r \mid F_{X^i}(r) - F_{X'^i}(r) \mid,
\end{equation*}
where $\sup_r$ is the supremum function \cite{rabanser2019failing}.

We compute an approximate $p$-value using the Kolmogorov-Smirnov distributions, with test statistic $\text{KS}$ and sample sizes $T$ and $T'$. The KS test assumes the samples are i.i.d., the measurement scale is at least ordinal, and $F(r)$ is continuous (though the test is more conservative if $F(r)$ is not continuous).

\paragraph{Chi-squared}
We use a Pearson's chi-squared test to evaluate the heterogeneity between firing rate counts of task and free-running data. To run the test for a given neuron $i$, we construct a contingency table with two rows and $R$ columns, where $R$ denotes the maximum firing rate observed in $X^i$ and $X^{'i}$. The first row contains the spike counts in task trials (i.e., $O_{1, r} = \sum_{i=1}^P \sum_{t=1}^T \mathbf{I}(X_t^i = r)$), and the second row the spike counts in free-running trials (i.e., $O_{2, r} = \sum_{i=1}^P \sum_{t=1}^{T'} \mathbf{I}(X_t^{'i} = r)$). Under the null hypothesis that the spike firing frequencies are not different between task and free-running trials, the expected frequency for a given cell $ij$ is
\begin{equation*}
    E_{ij} = \frac{\sum_{i=1}^2 O_{ij} \cdot \sum_{j=1}^R O_{ij} }{\sum_{i=1}^2 \sum_{j=1}^R O_{ij}}.
\end{equation*}
The test statistic is defined as
\begin{equation*}
    \chi^2 = \sum_{i=1}^2 \sum_{j=1}^R \frac{(O_{ij} - E_{ij})^2}{E_{ij}},
\end{equation*}
which under the null hypothesis follows a $\chi^2$ distribution with $C-1$ degrees of freedom \cite{rabanser2019failing}. 

The chi-squared test assumes data are i.i.d., the data in the table are counts or frequencies (rather than a transformation of the data, such as percentages), the levels of variables are mutually exclusive, and there are sufficiently large expected cell counts \cite{mchugh2013chi}.

\subsubsection{Multivariate tests}
\label{sec:multivariate.HTs}
We consider three multivariate tests: Hotelling's $T^2$ to test for equality of the mean vectors between samples, and independence tests mean maximum discrepancy and distance correlation, adapted for use as two-sample tests of distributional equivalence. We adopt a similar approach to our univariate analysis, where we evaluate whether firing rates in task and free-running trials are drawn from the same distribution at each location, and set our global estimate of the $p$-value for this to the minimum across confounds (after correcting for multiple comparisons). As above, we define a hypothesis test, $\text{HT}_{MV}(X_1, X_2)$ that returns a $p$-value corresponding to the significance of the difference. Unlike $\text{HT}_{UV}$, the input samples to $\text{HT}_{MV}$, $X_1$ and $X_2$, are multivariate rather than univariate. We express this approach formally in Algorithm \ref{alg:multivariate.alternative.HT}.

\begin{algorithm*}[!htb]
\caption{Multivariate alternative hypothesis tests}\label{alg:multivariate.alternative.HT}
\hspace*{\algorithmicindent} \textbf{Input}: task firing rates $(X_t)_{t=1}^T$ and stratification values $(Z_t)_{t=1}^T$, free-running firing rates $(X')_{t=1}^{T'}$ and stratification values $(Z'_t)_{t=1}^{T'}$, stratification levels $C$, and hypothesis test $\text{HT}_{MV}$ \\
\hspace*{\algorithmicindent} \textbf{Output}: $p$-value for the difference in context using alternative hypothesis test $\text{HT}_{MV}$, with Bonferroni correction for multiple comparisons
\begin{algorithmic}[1]
\Procedure{MultivariateAlternativeHypothesisTest}{}
\State $\text{pvalues} = [ \, ]$ \Comment{Init list with uncorrected $p$-values. See \S\ref{sec:subroutines}}

\For{$c \in C$} \Comment{}
    \State $X^c = [ \, ]$
    \State $X^{ic} =  [ \, ]$
    
    \For{$t = 1, \dots, T$}
        \If{$Z_t = c$}
            \State $X^c \gets \text{append}(X^c, X_t)$ \Comment{task neurons, stratification level $c$}
        \EndIf
    \EndFor
    
    \For{$t = 1, \dots, T'$}
        \If{$Z'_t = c$}
            \State $X^{'c} \gets \text{append}(X^{'c}, X'_t)$ \Comment{free-running neurons, stratification level $c$}
        \EndIf
    \EndFor
    
    \State $\text{pvalues} \gets \text{append}(\text{pvalues}, \text{HT}_{MV}(X^c, X^{ic}))$
\EndFor
\State \textbf{return} $\text{length}(C) \times \min (\text{pvalues})$ \Comment{Bonferroni correction}
\EndProcedure
\end{algorithmic}
\end{algorithm*}

\paragraph{Hotelling's \textbf{\textit{$T^2$}}}
Hotelling's $T^2$ is a generalization of Student's $t$-test to multivariate data and is used to test whether the mean vectors of two populations are significantly different \cite{hotelling1992generalization}. We let $\overline{X}$ and $\overline{X'}$ denote the columnwise means of $X$ and $X'$ (i.e., $\overline{X} = \frac{1}{T} \sum_{t=1}^T X_t$ and $\overline{X'} = \frac{1}{T} \sum_{t=1}^{T'} X'_t$). We then define the pooled covariance matrix:
\begin{equation*}
    \hat{\Sigma} = \frac{(T-1) X^\top X + (T' - 1) X'^\top X'}{T + T' - 2},
\end{equation*}
which allows us to define the test statistic as follows:
\begin{equation*}
    T^2 (X, X') = \frac{T \cdot T'}{T + T'} (\overline{X} - \overline{X'})^\top \hat{\Sigma}^{-1} (\overline{X} - \overline{X'}).
\end{equation*}
The $T^2$ statistic can be transformed so it follows an $F$-distribution as follows:
\begin{equation*}
    \frac{T + T' - P - 1}{P(T + T' - 2)} T^2 \sim F(P, T + T' - P - 1).
\end{equation*}
Hotelling's $T^2$ makes the i.i.d. assumption, and the assumptions that both groups follows the multivariate normal distribution with equal covariance matrices. \cite{panda2019nonparametric}

\paragraph{Independence tests as two-sample tests}

An independence test is a method for testing the hypothesis that the joint distribution of two random vectors, $X$ and $Y$, is independent; that is,
\[
H_{0, independence}: F_{XY} = F_X F_Y.
\]

In \cite{panda2019nonparametric}, the authors develop a methodology for using tests of independence in order to test the null hypothesis that two samples, $X$ and $X'$, come from the same distribution.  If $X$ is an $T \times P$ matrix and $X'$ is an $T' \times P$ matrix, define the matrices
\[
W = \begin{bmatrix}
X\\
X'
\end{bmatrix}
\]
\[
% Y = \begin{bmatrix}
% 1_{n} & 0_n\\
% 0_{n'} & 1_{n'}
% \end{bmatrix}
Y = \begin{bmatrix}
0_{T}\\
1_{T'}
\end{bmatrix}
\]
where $0_{T}$ and $1_{T'}$ denote $m\times 1$ matrices of 0's or 1's, respectively. We can then use $W$ and $Y$ to test if the data (contained in $W$) has dependence on the sample it was drawn from (contained in $Y$). If it does, the $Y$ is informative and we conclude $X$ and $X'$ are drawn from different distributions.

\paragraph{Distance correlation} The first independence test we consider is distance correlation ($\text{DCorr}$), which is used measure to measure the linear and nonlinear association between pairs of random variables of arbitrary dimensions \cite{szekely2007measuring}. We define $D^{X} \in \mathbb{R}^{P \times P}$ and $D^{X'} \in \mathbb{R}^{P \times P}$ to be the distance matrices of $X$ and $X'$, respectively. We then define matrix $C^X \in \mathbb{R}^{P \times P}$ as follows:
\begin{equation*}
    C_{ij}^{X} = \begin{cases}
      D_{ij}^X - \frac{1}{P-2} \sum_{t=1}^P D_{it}^X - \frac{1}{P-2} \sum_{t=1}^n D_{ij}^X + \frac{1}{(n-1)(n-2)} \sum_{t=1}^P D_{tt}^X & i=j, \\
      0 & \text{otherwise},
    \end{cases}
\end{equation*}
and define $C^{X'}$ similarly. This allows us to derive an unbiased estimate of the distance covariance \cite{szekely2014partial}:
\begin{equation*}
    \text{DCov} (X, X') = \frac{1}{n-3} \text{tr} (C^X C^{x'}),
\end{equation*}
from which we obtain the distance correlation test statistic, $\text{DCorr}$,
\begin{equation*}
    \text{DCorr} (X, X') = \frac{\text{DCov} (X, X')}{\sqrt{\text{DCov} (X, X) \cdot \text{DCov} (X', X')}} \in [-1, 1].
\end{equation*}
Distance correlation assumes data are i.i.d., but makes no assumptions about the underlying distribution.
To test the distributional equivalence of $X, X'$, we apply distance correlation to compute the test statistic $\text{DCov}(W, Y)$ and obtain a p-value using the chi-squared approximation \cite{shen2022chi}.

\paragraph{Mean maximum discrepancy} Maximum mean discrepancy (MMD) is a multivariate two-sample test \cite{gretton2012kernel} of independence. The unbiased estimate of the squared MMD statistic, with Gaussian kernel metric $k(\bullet, \bullet)$ is,
\begin{align*}
    \text{MMD}(X, X') &= \frac{1}{T(T-1)} \sum_{i=1}^T \sum_{j \neq i}^{T} k(X_i, X_j) 
    + \frac{1}{T'(T'-1)} \sum_{i=1}^{T'} \sum_{j \neq i}^{T'} k(X'_i, X'_j) \\
    &- \frac{2}{T \cdot T'} \sum_{i=1}^T \sum_{j = i}^{T'} k(X_i, X'_j).
\end{align*}
Similarly to distance correlation, MMD assumes samples are i.i.d., and to test the distributional equivalence of $X, X'$, we compute the test statistic $\text{MMD}(W, Y)$ and obtain a p-value using the chi-squared approximation \cite{shen2022chi}.

\subsubsection{Implementation}
\label{sec:alt.HT.implementation}
We used SciPy \cite{2020SciPy-NMeth}, distributed under the BSD license, to run the $t$-test \cite{heiman2001understanding} and KS test \cite{hodges1958significance}. For the multivariate tests, we used Hyppo \cite{panda2019hyppo}, a multivariate hypothesis testing package released under the MIT license. In the case of $\text{Dcorr}$ and $\text{MMD}$, for which analytical $p$-values are not available, we use a chi-squared approximation to generate test statistics and $p$-values. This approach was selected over permutation tests because the large number of parameter settings and seeds we evaluate the alternative hypothesis tests on, as well as the number of samples and dimensionality of the input data, rendered them too computationally expensive. According to previous work, the chi-squared approximation exhibits similar power to permutation tests \cite{shen2022chi}, and on a small subset of simulations in which we compared the $p$-values and test statistics derives from the chi-squared approximation and permutation tests, the results were similar.

\subsubsection{Expanded experimental results}
In Table \ref{tab:results.full}, we expand on the results presented in Table \ref{tab:results} by including the logistic regression and SVM classifiers, in addition to the Poisson decoder. 

\begin{table*}[ht!]
\centering
\caption{Results of the testing procedure to detect context effects in location encoding, for each of 2 development and 10 held-out mouse sessions, with a different set of neurons per session. The values $\obar{\text{acc}}_{\text{same}}$ and $\obar{\text{acc}}_{\text{cross}}$ (the estimates of $\text{Acc}_{same}$ and $\text{Acc}_{cross}$) are the average performance of the four decoders in Figure~\ref{fig:comparisons}, across 400 random seeds. The following columns show the the $p$-value for our test statistic using $\text{VIF}=12$ and a VIF estimated from data for the Poisson decoder, logistic regression, and linear SVM.}

\resizebox{\textwidth}{!}{\begin{tabular}{@{\extracolsep{4pt}}lccccccccc@{}}
\\
\toprule
& & & & \multicolumn{6}{c}{$p$-values} \\
 \cmidrule{5-10} & & & & \multicolumn{2}{c}{Poisson decoder} &  \multicolumn{2}{c}{logistic regression} & \multicolumn{2}{c}{linear SVM}\\
 \cmidrule{5-6} \cmidrule{7-8} \cmidrule{9-10}
 \thead{mouse/sess.\\(\# neurons)}  & \thead{$\obar{\text{acc}}_{\text{same}}$\\ ($\bar{\sigma}_\text{same}$)} & \thead{$\obar{\text{acc}}_{\text{cross}}$\\ ($\bar{\sigma}_\text{cross}$)} &  \thead{est.\\VIF} & \thead{fixed\\VIF\\=12} & \thead{using\\est.\\VIF} & \thead{fixed\\VIF\\=12} & \thead{using\\est.\\VIF} & \thead{fixed\\VIF\\=12} & \thead{using\\est.\\VIF} \\
 \cmidrule{1-10}
   
37/1 (41)	& 0.75 (0.007)	& 0.60 (0.007)	& 48	& \textbf{1.72e-3} & 0.081	 & 1.90e-2 & 0.148   & 2.09e-2	 & 0.154 \\
37/2 (38)	& 0.76 (0.007)	& 0.64 (0.007)	& 42	& \textbf{5.41e-3} & 0.135	 & 7.58e-3 & 0.133   & 6.86e-3	 & 0.124 \\ \\

36/1 (72)	& 0.81 (0.007)	& 0.66 (0.009)	& 40	& \textbf{3.53e-3} & 0.091	 & 6.88e-3 & 0.118   & 5.48e-3	 & 0.114 \\
36/2 (96)	& 0.71 (0.009)	& 0.52 (0.009)	& 58	& \textbf{1.43e-3} & 0.126	 & 8.10e-3 & 0.112   & 1.52e-2	 & 0.140 \\
38/1 (52)	& 0.61 (0.009)	& 0.35 (0.009)	& 57	& \textbf{4.71e-5} & 0.042	 & 2.71e-4 & 0.062	& 3.19e-4	 & 0.063 \\
38/2 (51)	& 0.56 (0.009)	& 0.38 (0.009)	& 45	& \textbf{2.72e-3} & 0.158	 & 5.36e-3 & 0.190	& 4.91e-3    & 0.175 \\
39/1 (42)	& 0.63 (0.008)	& 0.48 (0.008)	& 53	& \textbf{3.15e-3} & 0.112	 & 1.29e-2 & 0.177	& 6.45e-3	 & 0.156 \\
40/1 (104)  & 0.75 (0.007)   & 0.65 (0.008)   & 56	& \textbf{3.63e-2} & 0.248	 & 9.73e-2 & 0.309   & 1.17e-1	 & 0.320 \\
43/1 (45)   & 0.84 (0.007)   & 0.78 (0.007)   & 39	& 1.13e-1          & 0.269	 & 1.03e-1 & 0.256	& 1.15e-1	 & 0.280 \\
44/1 (42)	& 0.70 (0.007)   & 0.61 (0.008)   & 50	& \textbf{4.78e-2} & 0.213	 & 1.65e-2 & 0.184	& 1.09e-2	 & 0.168 \\
46/1 (56)	& 0.81 (0.007)   & 0.73 (0.008)   & 49	& \textbf{4.41e-2} & 0.248	 & 1.30e-2 & 0.155	& 1.91e-2	 & 0.160 \\
46/2 (57)	& 0.76 (0.006)   & 0.63 (0.007)   & 56	& \textbf{1.61e-3} & 0.102	 & 7.24e-4 & 0.084	& 2.52e-4	 & 0.068 \\

  \bottomrule
\end{tabular}}
\label{tab:results.full}

\end{table*}

We also conduct a similar analysis to that of the Table \ref{tab:results} using the alternative hypothesis tests. As shown in Table \ref{tab:alt.HT.results}, the alternative hypothesis tests are far less conservative than our method on the experimental data.

\begin{table}[!htbp]
    \centering
   \caption{Results of the testing procedure to detect context effects using the alternative hypothesis tests, with stratification by location and movement direction.}
    \begin{tabular}{lcccccc}
    \\
      \toprule
        & \multicolumn{6}{c}{$p$-values} \\
        \cmidrule{2-7}
        \thead{mouse/sess.\\ (\# neurons)} & \thead{$t^2$} & \thead{$\text{KS}$} & \thead{$\chi^2$} & \thead{$T^2$} & \thead{$\text{DCorr}$} & \thead{$\text{MMD}$}  \\
        \cmidrule{1-7}
        37/1 (41)	& 0         & 0         & 0         & 0 & 0         & 0 \\
        37/2 (38)	& 0         & 0         & 0         & 0 & 0         & 0 \\
        \\
        36/1 (72)	& 0         & 2.00e-273 & 0         & 0 & 0         & 0 \\
        36/2 (96)	& 0         & 6.18e-289 & 0         & 0 & 3.62e-308 & 0 \\
        38/1 (52)	& 0         & 2.04e-268 & 0         & 0 & 0         & 0 \\
        38/2 (51)	& 0         & 3.50e-318 & 0         & 0 & 0         & 0 \\
        39/1 (42)	& 1.20e-311 & 3.98e-217 & 6.80e-298 & 0 & 5.59e-291 & 1.73e-282 \\
        40/1 (104)	& 0         & 1.74e-270 & 0         & 0 & 0         & 0 \\
        43/1 (45)	& 1.69e-257 & 2.97e-091 & 3.22e-253 & 0 & 4.63e-223 & 5.99e-223 \\
        44/1 (42)	& 7.92e-087 & 1.29e-052 & 5.27e-094 & 0 & 4.14e-109 & 3.14e-109 \\
        46/1 (56)	& 3.98e-266 & 7.84e-153 & 7.98e-268 & 0 & 1.04e-291 & 1.27e-290 \\
        46/2 (57)	& 0         & 0         & 0         & 0 & 0         & 0 \\
    \bottomrule
    \end{tabular}
    \label{tab:alt.HT.results}
\end{table}
    
\subsection{Experiments on Synthetic Data}
\label{sec:synthetic.experiments}
To test our method using a known ground truth, and study its performance under a variety of conditions, we carried out a number of experiments with synthesized data. We generate datasets using the following generative model:
\begin{enumerate}
% tc_variance = 0.04, # tuning curve variance,
% tc_min_mean = 0.27, # min. mean of tuning curve, (max mean = 1-min. mean)
% rw_drift = 0.001,   # random walk drift
% rw_sigma = 0.05,    # random walk step st dev
    
    \item \textbf{Initial Setup}: A linear maze is set up on a unit interval, with location 0 and 1 representing the base and arms of the T-maze respectively. A mouse is initialized at at the base of the maze and is expected to travel to the arms and return back to the base. Each repetition is considered a subdataset. We generate $n_{\text{subdatasets}}$ number of independent subdatasets as required by our method.
    
    \item \textbf{Neuron Tuning Curves}: There are 3 types of neurons in our simulation: those that fire randomly, those that are location sensitive, and those that are context-dependent location sensitive. The number of random, location-sensitive, and context-dependent neurons are denoted $n_\text{random}$, $n_\text{both}$, and $n_\text{context}$ respectively. Each neuron is defined by a scaled Beta shaped tuning curve with $\alpha$ and $\beta$ shape parameters and scale $s$. The location tuning curves used in simulations are shown in Figure~\ref{fig:sim.tuning.curves}. They peak at a precise location, but have a wide range of adjacent locations where they will still fire. Note that this is much more demanding than the typical Gaussian tuning curve, which would have much more location specificity.
    Neurons that fire randomly have  $\alpha = \beta = 1$ to specify a uniform firing rate distribution across the maze, and show no location specificity. Location sensitive neurons have tuning curves specified with a mean $\mu_{\beta}$ and variance $\sigma^2_{\beta}$. Context-dependent location sensitive neurons have the additional ability to become insensitive to location and fire randomly given the context $c$ of the trial. 
    
    By default, half of the context-dependent neurons are location-sensitive on task trials only, and the other half are location-sensitive on free-running trials only. That is, if $n_\text{task}$ denotes the number of task-sensitive neurons, and $n_\text{FR}$ the number of free-running neurons, we have that $n_\text{context} = n_\text{task} + n_\text{FR}$ and $n_\text{task} = n_\text{FR} = n_\text{context}/2$.
    
    Given location $Y_t$ at time $t$ and tuning curve $f_{i,c}(x)$ for the $i$th neuron in context $c$, the spike count for each neuron $X_{i,t}$ is sampled from the conditional distribution 

    \[ 
    X_{i,t} | Y_t = y \sim\
    \begin{cases} 
     \text{Poisson}\left(s f_{i,c}(y; \alpha_i, \beta_i)\right) & \text{for loc. sensitive} \\
     \text{Poisson}\left(s f_{i,c}(y; 1, 1)\right) & \text{for loc. insensitive} \\
 \end{cases}
    \]
    
    where $f(y; \alpha, \beta)$ is the the p.d.f. of a beta distribution.
        
    \item \textbf{Mouse trajectory}: The mouse follows a reflected Gaussian random walk process, where the drift velocity reverses whenever the mouse reaches the boundary of the interval. We bias the random walk by specifying the drift velocity $v_d$, and variance $\sigma_{\text{walk}}^2$. At each time, we increment the location by a step size $\epsilon$ sampled from a $N(v_d,\sigma_{\text{walk}}^2)$ distribution, and then reflect if the location is greater or equal to the boundary at 1.  We end the walk when the mouse returns or moves past the initial location 0. Note that the mouse can and does repeatedly visit different locations before reaching the end of the maze and initiating the return, as illustrated by the sample trajectories in Figure~\ref{fig:sim.random.walk}.

    \begin{figure}[!htb]
        \centering
        \includegraphics[width=\linewidth]{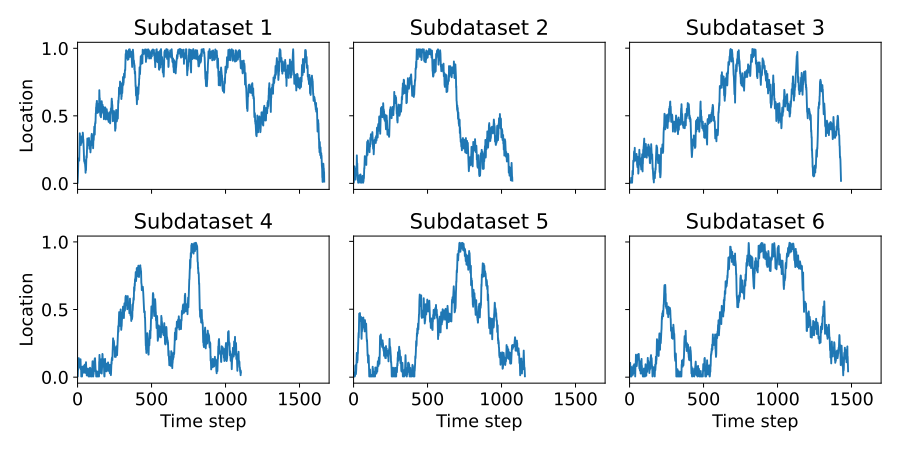}
        \caption{We show random walks from 6 generated datasets. Once the mouse travels from location 0 to 1, drift velocity reverses and the mouse travels back to location 0.}
        \label{fig:sim.random.walk}
    \end{figure}

    % \item The mouse is located in a unit interval [0,1].  The time points are discrete $t=1,\hdots, T$
    % \item The mouse follows reflected Gaussian random walk process where the drift velocity reverses whenever the mouse reaches the boundary of the interval.
    % \item The neurons have a context-dependent scaled Beta shaped tuning curves.  The context changes the scale of the tuning curve, or may turn the neuron ``off'', in which has the tuning curve becomes flat. Let $X(t)$ be the location of the mouse at time $t$, and let $f_{i,c}(x)$ be the tuning curve of the $i$th neuron at for the context $c$.   If the firing at time $t$ by neuron $i$ is written as $Y_i(t)$, then its conditional distribution is given by
    % \[
    % Y_i(t) \sim \text{Poisson}(f_{i,c}(X(t))).
    % \]
\end{enumerate}

The following simulation characteristics are matched with typical values from our real data (displayed in ~Table \ref{tab:dataset_specs}):
\begin{itemize}
    
    \item \textbf{Number of subdatasets}: For both Task and FR subdatasets, we set $n_{\text{subdatasets}} = 10$ -- approximately the total number of subdatasets in each session from Table \ref{tab:dataset_specs}.
    
    \item \textbf{Precision of the tuning curves}: We set the tuning curve variance to $\sigma^2_{\beta} = 0.01$ and select $\mu_{\beta}$ by evenly spacing out the total number of location sensitive neurons of a particular type on the interval $[0.15, 0.85]$. Representative tuning curves using these parameters are shown in Figure \ref{fig:sim.tuning.curves}.
    
    \begin{figure}[!htb]
        \centering
        \includegraphics[width=0.7\linewidth]{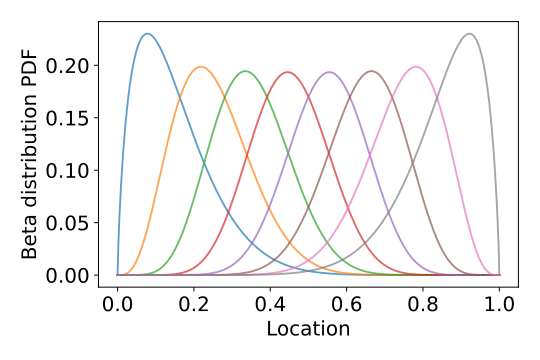}
        \caption{Location-sensitive tuning curves used in simulations. Here, we show 8 tuning curves, defined by beta distributions, evenly spaced across the maze. The universal tuning curve scale in this plot is 0.05, approximately equal to the median average firing of neurons in the experimental data.}
        \label{fig:sim.tuning.curves}
    \end{figure}

    \item \textbf{Number of time points}: We set the random walk parameters to $v_d = 0.001$ and $\sigma_{\text{walk}} = 0.03$. This gives us subdatasets with approximately 1000 timepoints each, approximately matching the number of time points in each trial of the experimental data.
\end{itemize}

\begin{figure}[!htb]
    \centering
    \includegraphics[width=\linewidth]{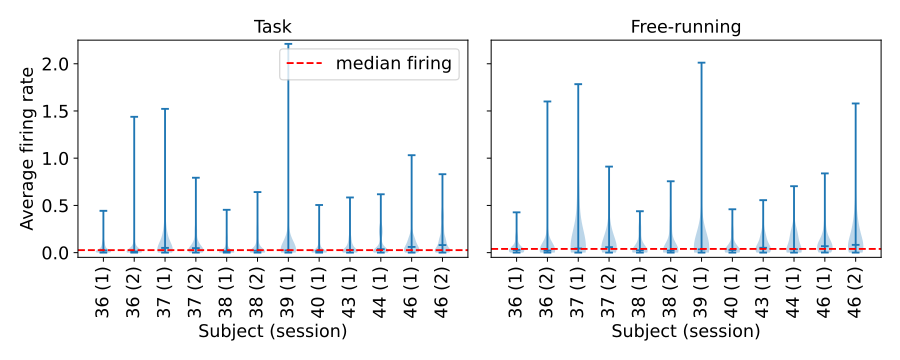}
    \caption{Firing rate patterns in the experimental data. For each subject, we calculate the average firing of each neuron in both task and free-running trials. The average firing rate of each neurons in each subject are shown in the violin plots. The median firing rate across all subjects and all neurons in task and free running trials is shown by the red dashed line.}
    \label{fig:experimental.firing.rates}
\end{figure}

In the simulations, we modulate the universal tuning curve scale factor ("scale") and the number and type of neurons. Here, scale denotes the mean firing of the neurons in the simulation; for a given neuron with scale $s$, the mean of the tuning curve PDF is $s$. This scale factor controls the spike count of each neuron, where higher scales correspond to higher spike counts and less noisy data. Our choice of scale factors in the simulations is informed by the firing rates in the experimental data. As shown in Figure \ref{fig:experimental.firing.rates}, the median average firing rate across task and free-running trials is approximately $0.05$, but there exists wide variation in the distribution of firing rates within and between subjects. For this reason, we evaluate the tests across a range of scales, $s = \{ 0.05, 0.20, 0.50, 2.00 \}$.

We conduct two analyses using the simulation data, both of which involve adjusting the number and type of neurons. In the first, our analysis of type I error shown in Figure \ref{fig:sim.type1.error}, we set $n_\text{random}$ and $n_\text{context}$ (i.e., the number of neurons that fire randomly and the number of neurons that are context-dependent location sensitive) to $0$, and adjust $n_\text{both}$ (i.e., the number of location-sensitive neurons that are not context-dependent) to values on the range $\{2, 3, ..., 50 \}$ with scales $\{ 0.05, 0.20, 0.50, 2.00 \}$. We run simulations using 100 random seeds for each parameter setting, and record the proportion of tests that report a $p$-value of $p \leq \alpha = 0.05$. Since the number of location-sensitive, context-dependent neurons is fixed at $0$, there is no difference in encoding so this proportion is the type I error rate for that setting.

In the second analysis, we investigate the power of our approach and compare it with the alternative hypothesis tests. We do so by adjusting $n_\text{random}$, $n_\text{both}$, and $n_\text{context}$, while holding the total number of neurons constant at 50 (approximately equal to the median number of neurons in the experimental data). For a given set of $n_\text{random}$, $n_\text{both}$, and $n_\text{context}$, we evaluate each hypothesis test on 100 random seeds and record the proportion of seeds for which the test rejects the null hypothesis, denoted $\Pr(\text{reject } H_0)$ (i.e., the proportion of seeds for which the hypothesis test found a significant difference in encoding between contexts). To compare the tests' performance across a range of environments, we first define the total signal, $n_\text{signal}$, in a given simulation as the number of location-sensitive neurons (both context-independent and context-dependent). That is, $n_\text{signal}=n_\text{both} + n_\text{context}$. Then, for a fixed total signal, we vary $n_\text{both}$ on the range $\{ 0, 2, ..., n_\text{signal} \}$ (and vary the corresponding $n_\text{context}$ such that $n_\text{context} = n_\text{signal} - n_\text{both}$). Since the total number of neurons is $50$, $n_\text{random} = 50 - n_\text{signal}$ for a given $n_\text{signal}$. Using this formulation, $\Pr(\text{reject } H_0)$ is the type I error rate at $n_\text{context} = 0$ (i.e., when there are no context-dependent neurons and therefore no difference in encoding between contexts), and the power at $n_\text{context} \geq 1$.

\begin{figure*}[!htb]
     \centering
    \subfloat[$\text{Scale} =0.05$]{%
      \includegraphics[width=\linewidth]{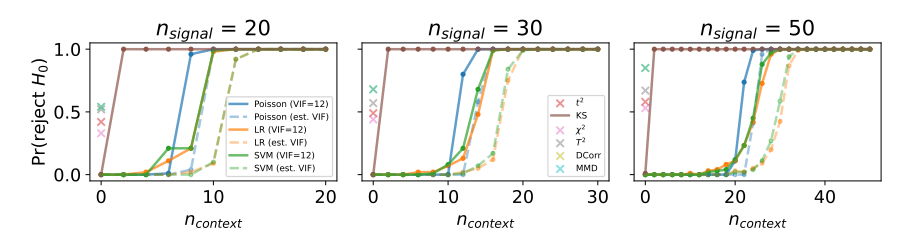}%
    }
    
    \subfloat[$\text{Scale} =0.20$]{%
      \includegraphics[width=\linewidth]{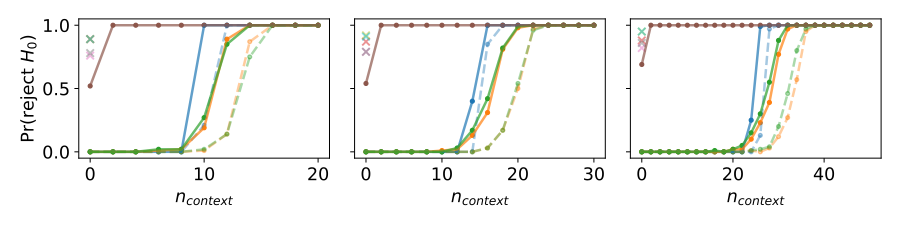}%
    }
    
    \subfloat[$\text{Scale} =0.50$]{%
      \includegraphics[width=\linewidth]{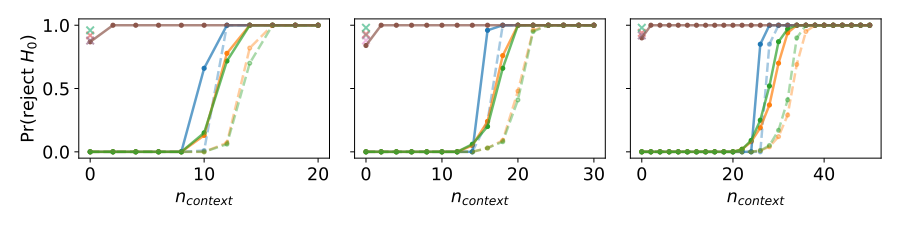}%
    }
    
        \subfloat[$\text{Scale} =2.00$]{%
      \includegraphics[width=\linewidth]{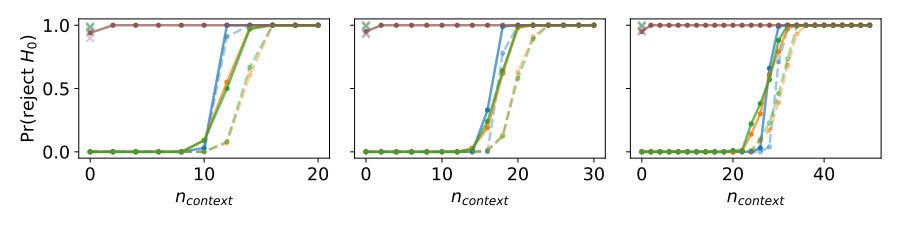}%
    }
    
    \caption{$\Pr(\text{reject } H_0)$ across different total signals (horizontally) and scales (vertically). For hypothesis tests with a high type I error rate at any scale (i.e., all of the alternative hypothesis tests \emph{except} Kolmogorov-Smirnov), we plot a single 
    value (using the \texttt{x} marker) at $n_\text{context}=0$ (i.e., the type I error) and omit values at $n_\text{context} > 0$ to improve legibility. We break up the legend and show it in two plots rather than one, also to maintain legibility. {\bf a, b, c, d:} Probability of rejecting the null hypothesis that there is no difference between distributions at a scale of (a) 0.05, (b) 0.20, (c) 0.50, (d) 2.00.}
    \label{fig:sim.varied.signal}
\end{figure*}

We conduct this evaluation of $\Pr(\text{reject } H_0)$ on the parameter grid $n_\text{signal} = \{20, 30, 50 \}, \text{scale} = \{0.05, 0.20, 0.50, 2.00 \}$, as shown in Figure \ref{fig:sim.varied.signal}. As in Figure \ref{fig:sim.type1.error}, as the scale and total signal increase, so does the type I error of the alternative hypothesis tests, while the symmetric difference remains conservative with a type I error rate of $0$. As the scale and total signal increase, the power of the symmetric difference (i.e., the number of context-dependent neurons to reach a given $\Pr(\text{reject } H_0)$) decreases. This is because, holding $n_\text{context}$ fixed, an increase in signal implies an increase in $n_\text{both}$. As $n_\text{both}$ increases, the difference between in- and cross-context accuracies is attenuated; with more location-sensitive neurons, the classifier relies less on each individual neuron, so the change in encoding from context-dependent neurons has less effect on the multivariate, global cross-context classification accuracy. The Poisson decoder demonstrates slightly higher or similar power to the logistic regression and linear SVM classifiers across $n_\text{signal}$ and $\text{scale}$ values.

\begin{figure*}[!htb]
     \centering
     
    \subfloat[From Task Decoder\label{fig:recovered.task.curves}]{%
      \includegraphics[width=\linewidth]{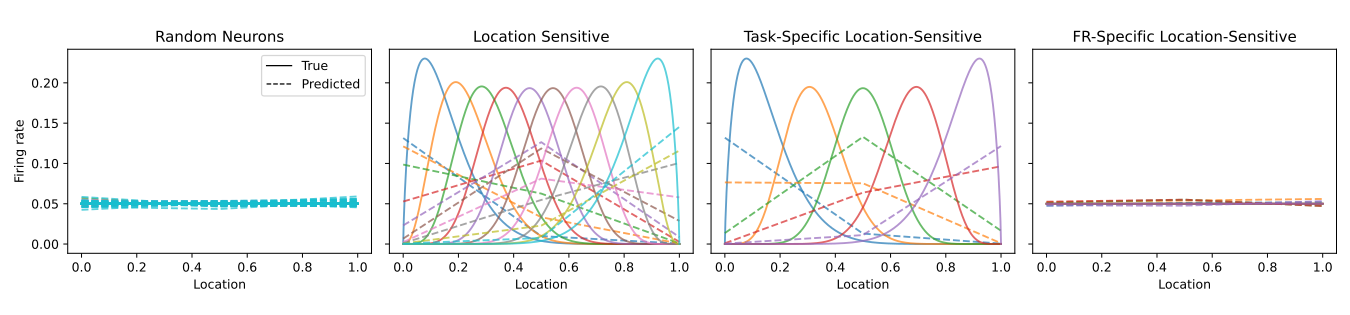} %
    }

    \subfloat[From FR Decoder\label{fig:recovered.fr.curves}]{%
    \includegraphics[width=\linewidth]{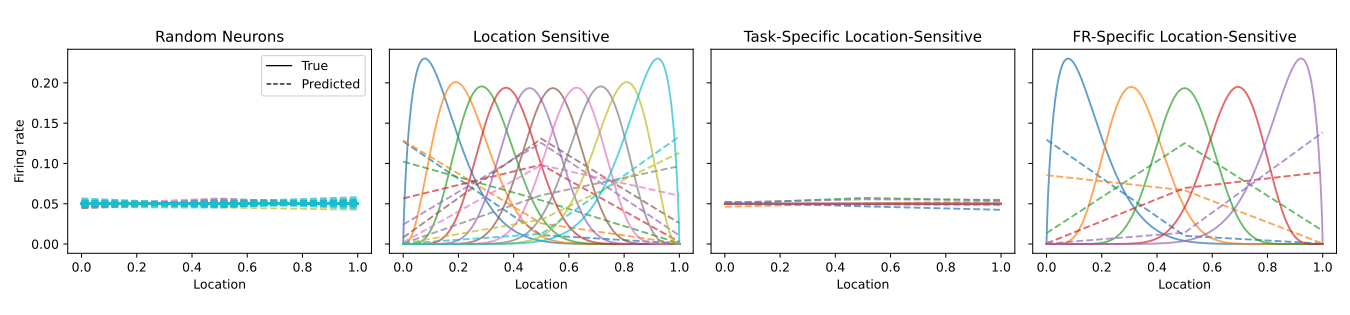} %
      
    }
    \caption{Recovered tuning curves from a simulation with multiple sessions, stochastic mouse trajectories for two different contexts (task vs. free-running.) The parameters are $n_\text{random} = 30$, $n_\text{both} = 10$, and $n_\text{context} = 10$ (i.e., $n_\text{task} = 5$, $n_\text{fr} = 5$). {\bf a, b:} Ground truth 1D location tuning curves (solid lines) and estimated tuning curves (dashed lines) recovered from Poisson decoder fits to the (a) task session synthetic data, or (b) free-running data. }
    \label{fig:recovered.tuning.curves}
\end{figure*}

Additionally, we illustrate the simulation ground truth tuning curves and demonstrate the recovery of the tuning curves by the Poisson decoder in Figure \ref{fig:recovered.tuning.curves}, for 50 neurons with a scale of $0.05$ and a total signal of $n_\text{power} = 20$.

\subsection{Subroutines used in algorithms}\label{sec:subroutines}

Please see Table \ref{tab:subroutines}.

\begin{table*}[!htbp]
    \centering
    \def\arraystretch{1.5}
    \begin{tabular}{cl}
        Subroutine & Description \\ 
        \hline
         permute($x$) & \multicolumn{1}{m{9cm}}{Takes in $x$, a list of values, and returns a list with the same values in a randomly shuffled sequence.} \\ 
         
         getCountsPerTargetClass($d$) & \multicolumn{1}{m{9cm}}{Takes in dataset $d$, and returns a list $[x_i], i=1,\dots, n_c$, where $x_i$ contains the counts of the $i$th target class.} \\
         
         draw($d$, $n$, $c$) & \multicolumn{1}{m{9cm}}{Randomly draws without replacement $n$ timepoints for class $c$ from dataset $d$. Returns a list of timepoints, $[t_i]$ for $i = 1, \dots, n$.} \\
         
         oversample($x$) & \multicolumn{1}{m{9cm}}{Takes in a list of timepoints, $x$, and returns $[t_i]$ for $i = 1, \dots, n_\text{maj}$, a list of timepoints where the minority classes are randomly oversampled to match the counts of the majority class, $n_\text{maj}$.} \\ 
         
         sort($x$) & \multicolumn{1}{m{9cm}}{Takes in a list of timepoints, $x$, and returns $[t_i]$ for $i = 1, \dots, \text{length}(x)$,  a list of timepoint arranged from least to greatest value across all sets.} \\
        
        append($Z$, $x$) & \multicolumn{1}{m{9cm}}{Takes in a list, $Z$, and appends $x$ to the list. If $Z =[ z_1, z_2, ..., z_n ]$, then $\text{append}(Z, x)$ returns $[z_1, z_2, ..., z_n, x]$. Note that $[\, ]$ denotes an empty list. That is, $\text{append}([\, ], x)$ returns $[x]$.} \\
    \end{tabular}
    \caption{Description of subroutines used in the algorithms}
    \label{tab:subroutines}
\end{table*}

\subsection{Computational complexity and runtime} \label{sec:computational.cost}
Our code is developed in Python 3.9.6, which is covered under the PFS license agreement. We create our plots using Matplotlib 3.4.2 \cite{Hunter:2007}, covered under the matplotlib license. The logistic regression and linear SVM are implemented using Scikit-learn \cite{scikit-learn}, under the BSD license.

We generate the experimental and simulation results using the National Institute of Health's Biowulf high performance computing cluster. In Table \ref{tab:total.runtime}, we specify the run times and total number of experimental runs to produce the results in Tables \ref{tab:results} and \ref{tab:results.full} and Figures \ref{fig:sim.type1.error} and \ref{fig:sim.varied.signal}. The total number of runs for the Poisson decoder is three times that of the logistic regression and linear SVM because we evaluated the Poisson decoder using three parameter settings (covariate matching and confound stratification, confound stratification only, and covariate matching only), but only considered covariate matching and confound stratification in the case of logistic regression and linear SVM.

\begin{table}[!htbp]
    \centering
  \caption{Average run time and total number of runs to produce primary results.}
    \begin{tabular}{clcc}
    \\
        \toprule 
        & \thead{Analysis} & \thead{Avg. time/run (sec.)} & \thead{Total runs} \\
        \cmidrule{1-4}
        \multirow{3}{*}{Real data (Tables \ref{tab:results}, \ref{tab:results.full})} & Poisson decoder & 198 & 14,400 \\
        & Logistic regression & 316 & 4,800 \\
        & SVM & 231 & 4,800 \\
        \cmidrule{1-4}
        \multirow{2}{*}{Simulation} & Type I error (Figure \ref{fig:sim.type1.error}) & 2,520 & 10,000 \\
        & $\Pr(\text{reject } H_0)$ (Figure \ref{fig:sim.varied.signal}) & 3,556 & 21,200 \\
    \bottomrule
    \end{tabular}
    \label{tab:total.runtime}
\end{table}

The bottleneck in our method is fitting the logistic regression, which has a cost that is dominated by a superlinear dependence on the dimensionality \cite{komarek2004logistic, minka2003comparison}. We evaluate the time complexity of the bottleneck using synthetic data for an increasing dimensionality and number of samples. We verify the time complexity to be close to linear by fitting a line on 100 points spaced across input sizes with a small order of magnitude and check to see if the linear trend holds when increasing the input size by several orders of magnitudes. To get a robust estimate of the linear trend, runtimes are repeated 5 times for a total of 500 points used for fitting. 

\begin{figure}[!htbp]
    \centering
    \includegraphics[width=\linewidth]{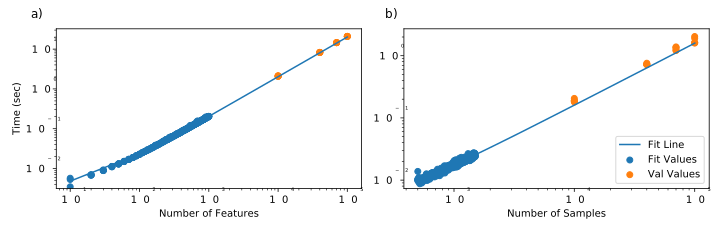}
    \caption{\textbf{a)} Runtime for increasing number of features up to $10^5$ attributes is very close to being linear. \textbf{b)} Runtime for increasing number of samples up to $10^5$ samples is slightly superlinear. }
    \label{fig:sim.runtime}
\end{figure}

When evaluating time complexity for the dimensionality, we fit a line on 100 values, evenly spaced from 10 to 1000. For the time complexity for the number of samples, we fit a line on 100 points, evenly spaced from 500 to 1500. For both analyses, we validate on 4 values evenly spaced from $10^4$ to $10^5$. The results are shown in Figure \ref{fig:sim.runtime}. The linear trend holds very well for an increasing dimensionality. For the number of samples, we see that the superlinear dependence is more obvious as the validation values tend to consistently lie above the fitted line.

% \begin{table*}[ht!]
% \centering
% \caption{Results of the testing procedure to detect context effects in location encoding, for each of 2 development and 10 held-out mouse sessions, with a different set of neurons per session. The values $\obar{\text{acc}}_{\text{same}}$ and $\obar{\text{acc}}_{\text{cross}}$ (the estimates of $\text{Acc}_{same}$ and $\text{Acc}_{cross}$) are the average performance of the four decoders in Table~\ref{fig:comparisons}, across 400 random seeds. The following columns show the the $p$-value for our test statistic using $\text{VIF}=12$, a VIF estimated from data, and the individual effects of not using VIF, covariate matching, and confound stratification.}

\subsection{Labeling Mouse Direction}\label{sec:fb.labeling}

\begin{figure*}[!htb]
     \centering
     
    \subfloat[Trajectory smoothing]{%
    %   \includesvg[width=0.7\linewidth]{recovered_task.svg}%
      \includegraphics[width=\linewidth]{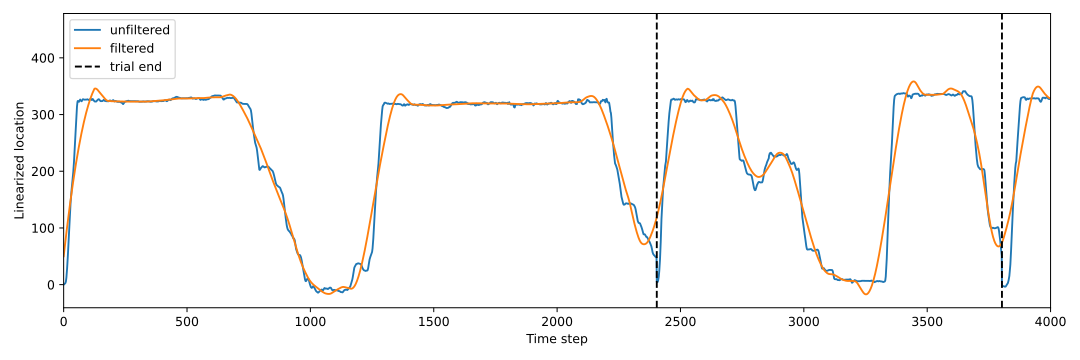} %
    }

    \subfloat[Direction labels]{%
    %   \includesvg[width=0.7\linewidth]{recovered_fr.svg}%
    \includegraphics[width=\linewidth]{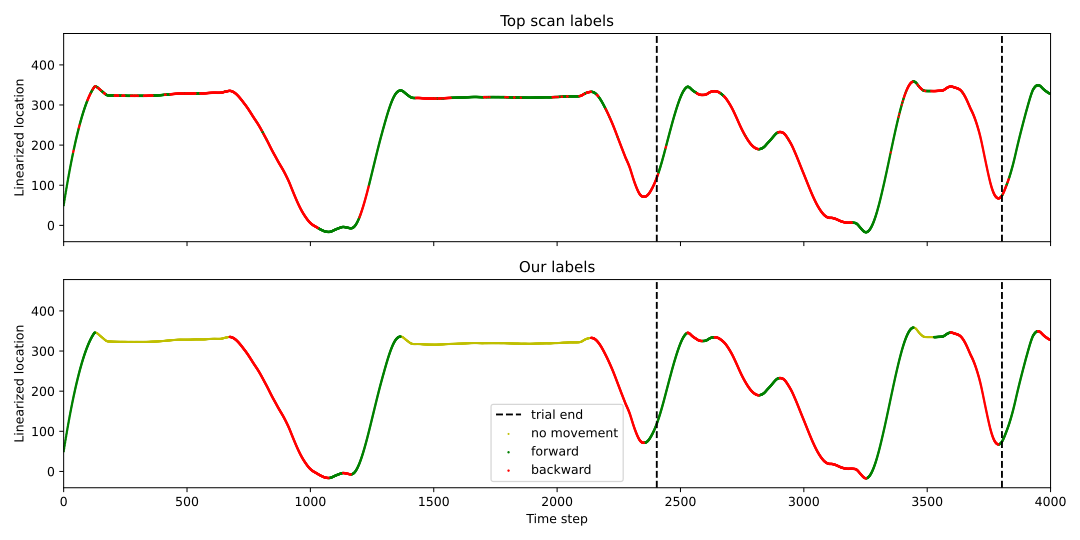} %
      
    }
    \caption{\textbf{(a)} The 4,000 time steps of location data from mouse 37 during its first series of task trials. Vertical dashed lines denote the end of the task trial. We use the smoothed trajectory to approximate the raw data. \textbf{(b)} Comparison of direction labels from the Topscan algorithm and our method. The Topscan algorithm coerces labels to be either F or B and does not identify regions where the mouse is not moving. Some regions (e.g., the region between time steps 1,500 and 2,000) have unstable labels, as direction labels oscillates between F and B. In our method, however, we correctly identify direction of movement and resolve regions with unstable labels.}
    \label{fig:direction.labels}
\end{figure*}

Head direction labels obtained from the overhead video footage using Topscan Suite 3.0 (Clever Sys) software was rough and inconsistent, particularly in regions where the mice had little to no movement. We develop our own method to get more robust labels for forwards and backwards directions. As shown in Figure \ref{fig:direction.labels}, we correctly identify regions of forwards and backwards movement as well as regions where the mouse has little or no movement. These direction labels are used for confound stratification in the main analysis.  We use only time points that are labeled as F or B and discard regions labeled as no movement. Our method, based on thresholding velocities on a smoothed trajectory, is outlined below.

\begin{enumerate}
    \item \textbf{Trajectory Smoothing}: We approximate the raw location data by smoothing out the mice trajectory using a Savitzky-Golay filter. This allows us to retain the majority of the movement features while eliminating much of the movement noise. We found that a window length of 251 time points (approximately 10 seconds of data) and a polynomial order of 2 gave us the best results for smoothing.

    \item \textbf{Velocity Thresholding}: 
    We obtain a preliminary direction label $d$ for each timepoint by setting a threshold $\tau$ for the velocity of the smoothed trajectory. The velocity $v$ is found by calculating the discrete first derivative for each value of the smoothed trajectory array. This allows us to detect movement that exceeds our predetermined velocity threshold. Timepoints are labeled according to the equation below. We found that a velocity threshold of 0.1 units per timepoint worked well for all sessions in our dataset. However, this value changes depending on the specifics of each experiment (i.e. the speed at which the animal is expected to travel, the resolution of the location units, the sampling rate).

    \[ 
    d(v) =\
    \begin{cases} 
      1 & v \textgreater \tau \\
      0 & -\tau \geq v \leq \tau \\
      -1 & v \textless -\tau 
   \end{cases}
    \]
    
    \item \textbf{Timeseries Segmentation}: Trajectory smoothing creates timeseries artifacts, such as artificial peaks and valleys as a result of a polynomial fit, especially in regions where the mouse transitions suddenly from fast to no movement, or vice versa. However, we take advantage of these artifactual peaks to help distinguish the different regions of movement. Using the peak detector, we identify the peaks and troughs to get segments of consecutive labels which will be used to filter our preliminary direction labels. We use the peak detector as implemented by Scikit-learn, specifying a prominence of 1.

    \item \textbf{Filtering Segments}: To avoid oscillating and unstable direction labels, we assume that each segment obtained above contains points which belong to the same direction label class. To reconcile mixed labels in each segment, we first quantify the amount of mixing by calculating the ratio of the minority to the majority direction label class counts. For ratios below 0.5, there is little mixing, and we simply relabel all values in the segment to match the majority direction class. For ratios above 0.5, there is much more mixing and we relabeling all timepoints in the segment according to the direction indicated by the mean velocity across the segment.
    
\end{enumerate}

%[YC: add algorithm?]
%[CZ: May not be necessary, since this is rather specific to our dataset and we will include the code anyways]

% \begin{figure}[!htbp]
%     \centering
%     \includesvg[width=\linewidth]{directionLabels2.svg}
%     \caption{\textbf{Top}: First 400 seconds of location data from mouse 37 during first free running session. We use the smoothed trajectory to approximate the raw data. \textbf{Middle}: Direction labels from Topscan algorithm. This method coerces labels to be either F or B and does not identify regions where the mouse is not moving. Some regions (i.e. region approximately from 35 to 50 seconds) have unstable labels as direction labels oscillates between F and B.  \textbf{Bottom}: Direction labels using our method. We correctly identify direction of movement and resolve regions with unstable labels.}
%     \label{fig:direction.labels}
% \end{figure}
\end{document}